\documentclass[10pt,twocolumn,letterpaper]{article}

\PassOptionsToPackage{table}{xcolor}

\usepackage[pagenumbers]{cvpr}

\definecolor{cvprblue}{rgb}{0.21,0.49,0.74}
\usepackage[pagebackref,breaklinks,colorlinks,allcolors=cvprblue]{hyperref}
\usepackage{amsmath}
\usepackage{graphicx}
\usepackage{booktabs}
\usepackage{multirow}
\usepackage{colortbl}
\usepackage{makecell}

\def\paperID{*****} 
\def\confName{CVPR}
\def\confYear{2026}

\title{CLEAR: Unlocking Generative Potential for Degraded Image Understanding in Unified Multimodal Models}

\author{
Xiangzhao Hao$^{1}$\thanks{Equal contribution.} \quad
Zefeng Zhang$^{2}$\footnotemark[1] \quad
Zhenyu Zhang$^{2}$\thanks{Project lead.} \quad
Linhao Yu$^{2}$ \quad
Yao Chen$^{2}$ \\
Yiqian Zhang$^{2}$ \quad
Haiyun Guo$^{1}$\thanks{Corresponding author.} \quad
Shuohuan Wang$^{2}$ \quad
Yu Sun$^{2}$ \\
$^{1}$Institute of Automation, Chinese Academy of Sciences \\
$^{2}$Baidu Inc. \\
{\tt\small haoxiangzhao2023@ia.ac.cn}
}

\begin{document}
\maketitle

\begin{abstract}
Image degradation from blur, noise, compression, and poor illumination severely undermines multimodal understanding in real-world settings. Unified multimodal models that combine understanding and generation within a single architecture are a natural fit for this challenge, as they understand clean images well and their generative pathway can model the fine-grained visual structure that degradation destroys. Yet when directly answering questions about degraded images, these models fail to leverage their own generative capacity. Generation and understanding coexist but remain functionally disconnected. We trace this disconnect to the fact that existing training regimes never ask the model to invoke generation during reasoning, and the standard decode-reencode pathway between the two capabilities does not support effective joint optimization. Together, these prevent answer-level feedback from shaping how the model generates. We present CLEAR, a framework that connects the two capabilities through three progressive steps: (1) supervised fine-tuning on a degradation-aware dataset to establish the generate-then-answer reasoning pattern; (2) a Latent Representation Bridge that replaces the decode-reencode detour with a direct, optimizable connection between generation and reasoning; (3) Interleaved GRPO, a reinforcement learning method that leverages this connection to jointly optimize text reasoning and visual generation under answer-correctness rewards. Freed from pixel-level regression targets, the model learns to generate intermediate visual states that not only serve downstream reasoning but also exhibit higher perceptual quality than those produced under explicit reconstruction supervision, revealing that task-driven optimization and visual quality are naturally aligned rather than in conflict. We construct MMD-Bench, covering three degradation severity levels across six standard multimodal benchmarks. Experiments show that CLEAR substantially improves robustness on degraded inputs while preserving strong clean-image performance, confirming that the generative and understanding capabilities within unified models can be effectively connected for robust visual understanding. Our code and data are publicly available at \url{https://github.com/haoxiangzhao12138/CLEAR}.
\end{abstract}

\begin{figure}[h]
    \centering
    \includegraphics[width=\linewidth]{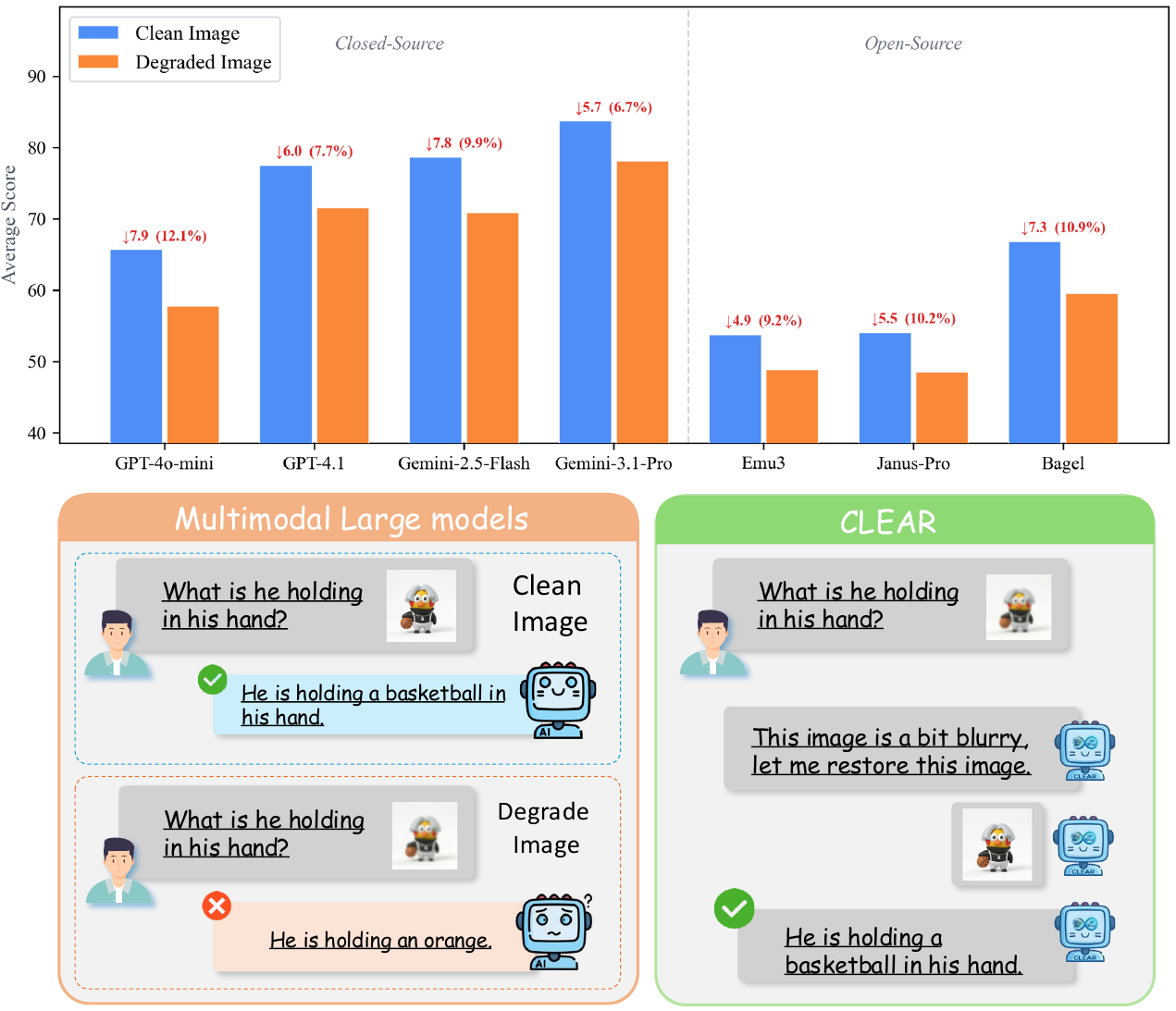}
\caption{Top: average scores of commercial and open-source multimodal models on clean versus degraded inputs from MMD-Bench across six benchmarks. All models show substantial performance drops under degradation. Bottom: comparison between existing multimodal models and CLEAR on a degraded image.}
    \label{fig:teaser}
\end{figure}

\section{Introduction}


Image degradation is a routine part of real-world visual data, not an edge case.
Images from autonomous driving, surveillance, mobile photography, and video conferencing are frequently corrupted by motion blur, sensor noise, poor illumination, and aggressive compression.
These degradations damage the low-level visual cues that multimodal models depend on for recognition, grounding, and reasoning~\cite{imagenetc, rbench}.
As Figure~\ref{fig:teaser} illustrates, multimodal models can correctly identify an object in a clean image yet misrecognize it entirely when the same image is degraded.
This is not an isolated failure.
Across commercial systems such as GPT-4o~\cite{gpt4o} and open-source architectures of varying scales~\cite{llava, qwen2vl, internvl, zhang2025cooper}, we observe substantial accuracy losses on degraded versions of six standard benchmarks, indicating that sensitivity to image degradation is a pervasive vulnerability across the current multimodal landscape.
Robustness to such degradations is a core requirement for deploying multimodal systems in practice.


Among existing architectures, unified multimodal models stand out for their ability to handle both visual understanding and image generation within a single model.
Rather than relying on separate specialist modules, these models share a common backbone across the two tasks~\cite{bagel, janus, emu3, chameleon, transfusion}, with a vision encoder~\cite{clip, siglip} that maps images into semantic features for understanding and a generative pathway that operates through a VAE~\cite{kingma2013auto, rombach2022high} or discrete tokenizer~\cite{esser2021taming} to produce images from continuous or quantized latent representations.
The understanding pathway excels at high-level semantic reasoning, including object recognition, spatial relationship inference, and visual question answering, when the input image is clean.
The generative pathway, by contrast, operates at a fundamentally different level of visual representation, capturing low-level structure such as texture, edge detail, color distribution, and spatial layout that high-level semantic features tend to discard~\cite{zhang2023tale}.
Degraded image understanding aims to enable models to interpret images whose low-level visual cues have been unintentionally corrupted, and to answer questions about the high-level semantic information they contain. 
In unified models, the understanding and generation pathways naturally correspond to these two types of features: the former primarily captures high-level semantics, while the latter models low-level visual details.

Yet when asked to directly answer questions about degraded images, unified models fail to bring these two capabilities together.
As the top panel of Figure~\ref{fig:teaser} shows, Bagel~\cite{bagel}, Janus-Pro~\cite{janus}, and Emu3~\cite{emu3} all suffer substantial performance drops under degradation, with no sign that their generative pathway contributes to robustness.
The model does not spontaneously invoke generation to compensate for the visual information that degradation has destroyed.
Generation and understanding coexist in the same architecture but remain functionally disconnected.
This motivates the central research question of this work: how can we connect  generation with the reasoning process to support understanding on degraded images?

To answer this question, we attribute the disconnect to two compounding factors.
\textbf{(1) Behavioral:} Existing unified models are never trained to invoke generation as part of the reasoning process for understanding tasks.
Their training treats generation and understanding as separate objectives, so the model has no experience with a reasoning pattern that uses generated visual content to support answer production.
\textbf{(2) Structural:} Even if such a pattern were introduced, the standard pathway connecting generation to understanding requires that the generated latent representations be decoded into pixel space and re-encoded through a frozen vision encoder before they can influence reasoning.
The frozen decoder and encoder sever the computation graph between the generation and understanding stages, preventing gradients from answer-level supervision from propagating back to the parameters that control what the model generates.
Taken together, the two factors reinforce each other.
Without the behavioral pattern, the model never attempts to generate for understanding, and the structural bottleneck is never even exposed.
Without a differentiable optimization route, introducing the behavioral pattern alone cannot teach the model what to generate, only that it should.


To bridge this disconnect, we propose CLEAR (\textbf{C}omprehension via \textbf{L}atent \textbf{E}nhancement and \textbf{A}daptive \textbf{R}easoning), a framework that connects the generative and understanding capabilities of unified models through three progressive steps.
\textbf{(1) Behavioral Initialization.}
We construct a degradation-aware training dataset where samples with mild or no degradation receive direct-answer supervision and samples with severe degradation require the model to first generate an intermediate visual state before answering.
Fine-tuning on this dataset teaches the model the generate-then-answer reasoning pattern and establishes when to invoke generation and how to structure the interleaved trajectory.
\textbf{(2) Latent Representation Bridge.}
With the behavioral pattern in place, the next bottleneck is the decode-reencode pathway.
CLEAR addresses this by injecting generated latent representations directly into the reasoning context, eliminating the pixel-space detour entirely.
This allows generated visual information to participate in reasoning alongside the original encoded features.
At the same time, it creates a direct, differentiable connection from generation to reasoning that makes the joint training in the next step possible.
\textbf{(3) Interleaved GRPO.}
With the bridge providing an effective optimization route, we apply Interleaved GRPO, a reinforcement learning method building on GRPO~\cite{grpo} and Flow-GRPO~\cite{flowgrpo} that jointly optimizes text reasoning and visual generation within a shared forward pass.
The reward centers on final answer correctness, so answer-level feedback now flows through the bridge to shape how the model generates.
Within this training, the model also learns an adaptive generation strategy that evaluates input quality during reasoning and invokes generation only when degradation is likely to impair understanding, avoiding unnecessary computation on clean inputs.


For evaluation, we construct MMD-Bench by applying 16 real-world corruption types at three severity levels to 6 widely used multimodal benchmarks, and additionally evaluate on R-Bench~\cite{rbench}, an existing benchmark for degraded-image understanding.
Experiments show that CLEAR substantially improves degraded image understanding while maintaining strong clean-image performance.
Our analysis further reveals a finding that may seem counter-intuitive.
When pixel-level reconstruction supervision is removed and only answer-correctness rewards remain, the model not only preserves but even improves the perceptual quality of its generated intermediate states. 
This suggests that visual quality is naturally aligned with task optimization, and that explicit reconstruction supervision is more a constraint than a requirement.
Our main contributions are as follows.
\begin{itemize}
    \item We identify a functional disconnect in unified multimodal models where generation and understanding coexist but fail to cooperate under degraded inputs. To address this, we construct a degradation-aware training set with difficulty-dependent supervision that teaches unified models to invoke generation as part of the reasoning process.
    \item We propose CLEAR, which bridges this disconnect through Behavioral Initialization via supervised fine-tuning, a Latent Representation Bridge that opens a direct optimization route from generation to reasoning, and Interleaved GRPO that jointly optimizes understanding and generation with answer-correctness rewards.
    \item 
    Experiments on MMD-Bench and R-bench confirm that CLEAR achieves substantial robustness gains on degraded inputs without sacrificing clean-image performance. Our analysis further shows that removing pixel-level supervision  leads to intermediate visual states with higher perceptual quality, indicating that task-driven optimization can naturally aligns with visual quality.
\end{itemize}

\section{Related Work}

\textbf{Robustness under Image Degradation.} The vulnerability of visual recognition systems to low-level image degradations has been studied extensively since ImageNet-C~\cite{imagenetc}, which showed that modern classifiers suffer substantial accuracy drops under blur, noise, weather effects, and digital distortions.
This line of work has since been extended to the multimodal setting.
Several benchmarks have been proposed to evaluate vision-language models under degraded conditions, revealing that even models built on strong visual encoders such as CLIP~\cite{clip} remain highly sensitive to degraded inputs in tasks including visual question answering, captioning, and multimodal reasoning~\cite{mmebench, rbench, zhao2024evaluating}.
As we demonstrate in Section~\ref{sec:experiments}, unified multimodal models such as Bagel~\cite{bagel}, Janus-Pro~\cite{janus}, and Emu3~\cite{emu3} suffer comparable drops, indicating that their generative pathways do not spontaneously contribute to robustness.
While existing efforts have explored corruption-aware data augmentation~\cite{hendrycks2020augmix, mintun2021interaction} and external restoration pipelines~\cite{swinir, restormer, nafnet} to mitigate degradation effects, unified models already possess a generative pathway that operates on exactly the low-level visual structure that degradation destroys, yet this internal capacity is never activated during understanding tasks.
Our work focuses on how to connect this capacity to the understanding process so that the model can compensate for degradation from within.

\noindent\textbf{Unified Vision-Language Models.}
Recent work has moved toward unifying visual understanding and image generation within a single model architecture.
Systems such as Chameleon~\cite{chameleon} and Emu3~\cite{emu3} adopt discrete visual tokenization through vector quantization, representing images and text in a shared token space for autoregressive generation of interleaved multimodal sequences.
More recent models including Janus~\cite{janus}, Bagel~\cite{bagel}, and Transfusion~\cite{transfusion} operate on continuous latent representations through variational autoencoders, which preserve richer low-level visual information compared to discrete tokens.
Other representative systems such as VILA-U~\cite{vila-u}, Show-o~\cite{show-o}, and Unified-IO~2~\cite{unifiedio2} explore different trade-offs between generation quality and understanding performance.
A distinctive property of unified architectures is their ability to interleave text generation with image generation, opening the possibility for richer reasoning trajectories than understanding-only models can support.
However, this potential remains largely unexplored for visual understanding under degraded conditions.
In most current pipelines, generated visual content must be decoded into pixel space and re-encoded by the vision encoder before it can influence subsequent reasoning steps, a procedure that is both computationally expensive and unfavorable for joint optimization.
How to effectively route generative representations into the understanding pipeline so that generation actively supports reasoning is the question our work aims to address.

\noindent\textbf{Reinforcement Learning for Vision-Language Reasoning.}
Reinforcement learning has emerged as a powerful approach for improving reasoning capabilities beyond what supervised fine-tuning can achieve.
Methods such as GRPO~\cite{grpo} and DeepSeek-R1~\cite{deepseekr1} optimize directly for outcome-level rewards, enabling models to discover effective reasoning strategies for mathematical and logical problems without step-level supervision.
Recent efforts have begun extending RL to multimodal reasoning, using rule-based or outcome-level rewards to improve visual question answering and grounding~\cite{visionr1, vlmr1, r1v}.
In parallel, diffusion-based or flow-based policy optimization methods such as DDPO~\cite{ddpo}, DiffusionDPO~\cite{diffusiondpo}, and Flow-GRPO~\cite{flowgrpo} apply RL to visual generation under learned reward signals.
Despite this progress, existing methods optimize text and image generation in isolation.
Unified multimodal models introduce a fundamentally different setting where text and image outputs form a single interleaved trajectory, and the value of generated visual content should be judged by how much it contributes to the final reasoning outcome rather than by its standalone appearance.
This requires coordinated optimization where both modalities share a computation graph under a common end-task reward.
Our Interleaved GRPO addresses this gap by jointly optimizing text reasoning and visual generation in a single forward pass, with reward centered on answer correctness.

\begin{figure*}[t]
    \centering
    \includegraphics[width=0.8\textwidth]{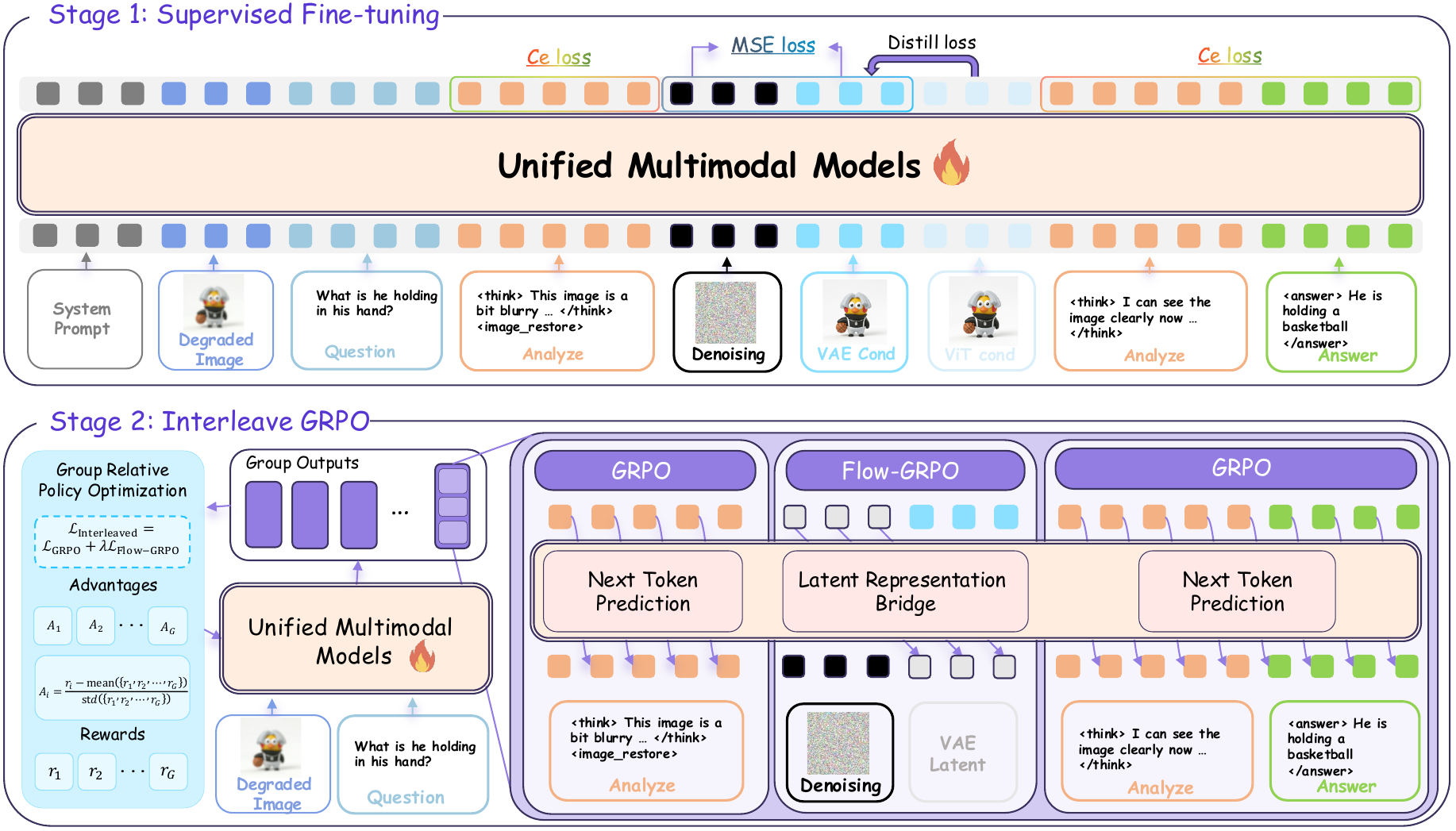}
\caption{Overview of CLEAR. Stage~1 (top) performs supervised fine-tuning to establish the generate-then-answer reasoning pattern and warm-start the Latent Representation Bridge, with both VAE latent and ViT re-encoded features injected during this stage. Stage~2 (bottom) applies Interleaved GRPO, where text tokens are optimized with GRPO and the denoising step with Flow-GRPO, sharing the same group-relative advantage from answer-correctness rewards. The ViT path is removed in Stage~2, making the bridge the sole connection between generation and reasoning.}
    \label{fig:pipeline}
\end{figure*}


\section{Method}

\subsection{Overview}
\label{sec:overview}

\textbf{Base Architecture.}
CLEAR is built on Bagel-7B~\cite{bagel}, a unified vision-language model that supports both understanding and generation within a shared Mixture-of-Transformer backbone.
The understanding pathway encodes images through a SigLIP~\cite{siglip} vision encoder, while the generation pathway operates through a VAE~\cite{kingma2013auto} that maps between pixel space and a continuous latent space.
Both pathways feed into the same language model, where generation and understanding share a common reasoning space.

\noindent\textbf{Training Pipeline.}
As shown in Figure~\ref{fig:pipeline}, CLEAR trains the model in two stages.
Stage~1 performs supervised fine-tuning on a degradation aware dataset to teach the model the generate-then-answer reasoning pattern and warm-start the Latent Representation Bridge so that the language model can begin reading information from the generated VAE latent.
Stage~2 applies Interleaved GRPO, which leverages the bridge as a differentiable connection to jointly optimize text reasoning and visual generation under answer-correctness rewards, during which the model also acquires an adaptive strategy that decides when generation is needed.

\noindent\textbf{Reasoning Trajectory.}
Given an input, the model first enters an analysis phase within \texttt{<think>}, where it reasons about the visual content and implicitly assesses whether generation would improve its answer.
If the model chooses to generate, it emits the \texttt{<image\_restore>}, which triggers multi-step denoising to produce an intermediate visual state in VAE latent space.
The resulting latent tokens are injected directly into the reasoning context through the Latent Representation Bridge, serving as the visual input for subsequent reasoning.
Rather than decoding to pixel space and re-encoding through a vision encoder, the model reasons directly over the generated latent representation, performing what we term \emph{latent reasoning}.
Another analysis phase then processes these latent tokens alongside the preceding text to produce the final answer within \texttt{<answer>}.
When the model judges that the available visual information is sufficient, it skips generation entirely and proceeds directly to the answer, keeping the trajectory compact.

\subsection{Behavioral Initialization through SFT}
\label{sec:sft}

The first step addresses the behavioral gap.
Existing unified models have never been trained to invoke generation as part of the reasoning process for understanding tasks.
We bridge this gap through supervised fine-tuning on a purpose-built degradation-aware dataset.

\textbf{Training Data Construction.}
We sample a subset from the LLaVA-OneVision~\cite{llavaonevision} instruction-tuning dataset.
For each sampled image, we apply degradations drawn from a pool of 16 corruption types covering four categories (capture, transmission, environment, and post-processing) at three intensity levels, and then evaluate whether the base Bagel model can correctly answer the associated question on the degraded version. The full list of corruption types and their severity parameters are provided in the supplementary material~\ref{app:data}.
Samples that the model answers correctly are assigned the direct-answer pathway, while samples it fails on are assigned the generate-then-answer pathway.
For both types, we use GPT-4.1~\cite{gpt41} to generate structured reasoning traces, with direct-answer traces containing analysis and answer phases and generate-then-answer traces additionally containing the generation trigger and post-generation analysis.
All traces are filtered against ground-truth answers to remove incorrect reasoning.
The final SFT dataset contains 24k samples, split evenly between the two pathway types.
A separate non-overlapping set of 24k samples is reserved for the Interleaved GRPO stage. Since LLaVA-OneVision is the same corpus used to train the base Bagel model, any potential overlap with evaluation benchmarks affects all compared methods equally.

\textbf{Training Objective.}
The core objective is next-token prediction over the text tokens in the trajectory ($\mathcal{L}_\text{CE}$), which teaches the model the interleaved reasoning format and the conditions under which generation should be triggered.
Two auxiliary losses support the visual generation side.
An MSE loss ($\mathcal{L}_\text{MSE}$) provides an initial training signal for the denoising process by encouraging the generated VAE latent to approximate the clean image in latent space.
A distillation loss ($\mathcal{L}_\text{KL}$) uses the ViT features of the clean image as the teacher signal to guide the VAE latent representations.
Since the language model has been pretrained exclusively with ViT features as visual input, raw VAE latent tokens fall outside the representation distribution it can interpret.
The KL loss addresses this by encouraging the VAE latent hidden states to move toward the ViT feature distribution at each transformer layer, with higher layers receiving greater weight.
This does not collapse the two representations into identical outputs.
Rather, it teaches the language model to read useful information from the VAE latent path while the VAE representations retain their characteristic low-level structural content that ViT features lack.
The overall objective is
\begin{equation}
    \mathcal{L}_\text{SFT} = \mathcal{L}_\text{CE} + \lambda_\text{MSE}\,\mathcal{L}_\text{MSE} + \lambda_\text{KL}\,\mathcal{L}_\text{KL}.
\end{equation}
During SFT, both the VAE latent and the ViT re-encoded features of the generated image are injected into the reasoning context after denoising.
The ViT re-encoded features serve as an auxiliary input that supports the KL distillation loss and provides the model with a familiar representation format during the early stages of bridge training; they are removed in the GRPO stage once the bridge is established (Section~\ref{sec:grpo}).
The SigLIP vision encoder and VAE encoder/decoder remain frozen throughout.
Only the language model backbone is updated.

After SFT, the model has learned when to generate and how to structure the interleaved trajectory, but what it generates remains constrained by the MSE target.
While the clean-image latent provides a reasonable initialization for the denoising process, the MSE objective suffers from a well-known regression-to-mean tendency~\cite{mathieu2016deep, ledig2017photo} that limits the sharpness and perceptual quality of the generated states.
To move beyond this ceiling, the model needs a training signal that connects generation directly to answer correctness, which is what the next two steps provide.

\begin{figure}[t]
    \centering
    \includegraphics[width=\linewidth]{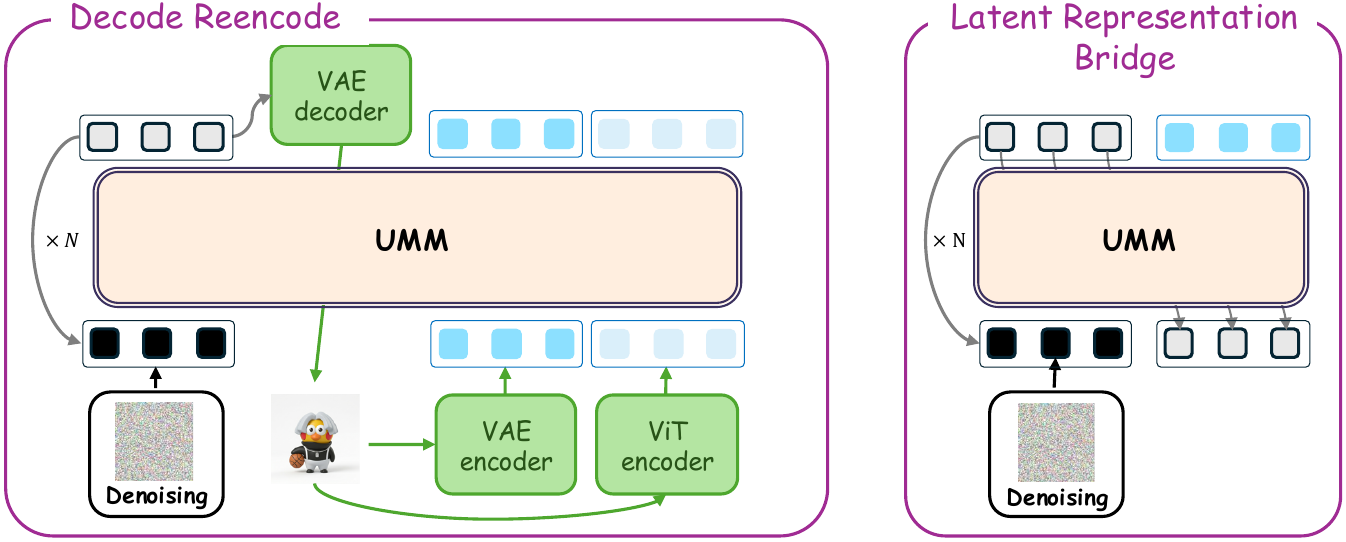}
    \caption{Left: the standard decode-reencode path in existing unified models. The generated VAE latent must be decoded into pixels and re-encoded through the ViT before it can enter the reasoning context. Right: the Latent Representation Bridge in CLEAR. The generated VAE latent is directly concatenated into the reasoning context, eliminating the decode-reencode bottleneck and providing an effective optimization route from answer correctness back to generation.}
    \label{fig:bridge}
\end{figure}

\subsection{Latent Representation Bridge}
\label{sec:bridge}

The second step addresses the structural barrier that prevents generation from being jointly optimized with understanding.

As illustrated in Figure~\ref{fig:bridge} (left), existing unified models route generated visual content through a lengthy detour before it can participate in reasoning.
The VAE latent produced by the denoising process must first be decoded into pixel space, then re-encoded through the vision encoder, before the resulting features can enter the language model context.
This path adds substantial computational cost and, more importantly, severs the gradient connection between generation and reasoning, because the frozen decoder and encoder sit between the two stages and block backpropagation.

CLEAR replaces this detour with a direct connection, as shown in Figure~\ref{fig:bridge} (right).
After denoising, the generated VAE latent tokens are concatenated into the reasoning context alongside the original ViT features and text tokens.
This gives the language model two complementary sources of visual evidence for reasoning: high-level semantic information from the ViT features of the degraded input and fine-grained structural detail from the generated VAE latent.

The more critical consequence is for training.
Because the generated latent now participates directly in the computation that produces the answer, answer-level supervision can reach the generation process through a differentiable path.
This is what makes joint optimization in the next step possible.
During SFT, the KL distillation loss has already provided a warm start for this connection so that the language model can begin exploiting information from the VAE latent tokens.
In the GRPO stage that follows, the ViT re-encoding route used during SFT is removed, and the bridge becomes the sole connection between generation and reasoning.
This ensures that answer-correctness rewards flow entirely through the bridge, freeing the generation process from pixel-level regression targets and allowing it to be shaped by downstream task performance.

\begin{table*}[t]
\centering
\caption{Main results under Hard degradation. R-Bench-Dis is an existing degraded-image benchmark; the remaining six are from MMD-Bench. Best in \textbf{bold}, second best \underline{underlined}. $^\dagger$Closed-source results are included as reference points and are not directly comparable due to differences in model scale and training data.}
\label{tab:main}
\resizebox{0.8\textwidth}{!}{
\begin{tabular}{l|cccccc|c|c}
\toprule
& \multicolumn{6}{c|}{MMD-Bench (Hard)} & & \\
\cmidrule(lr){2-7}
Method & MMBench & MM-Vet & MMVP & CV-Bench & MMStar & RealWorldQA & R-Bench-Dis & AVG \\
\midrule
\multicolumn{9}{l}{\textit{Closed-source models$^\dagger$}} \\
GPT-4o-mini & 67.02 & 50.91 & 64.00 & 59.87 & 45.93 & 58.95 & 61.21 & 58.27 \\
GPT-4.1-mini & 76.08 & 51.88 & 71.00 & 74.96 & 60.73 & 72.41 & 72.52 & 68.51 \\
Gemini-2.5-Flash & 79.33 & 66.55 & 72.33 & 76.01 & 62.00 & 69.15 & 72.72 & 71.16 \\
\midrule
\multicolumn{9}{l}{\textit{Open-source unified models}} \\
Emu3 & 53.71 & 21.51 & 65.00 & 58.34 & 42.06 & 52.55 & 55.15 & 49.76 \\
Janus-Pro & 55.57 & 31.33 & 52.66 & 66.75 & 41.53 & 43.52 & 49.09 & 48.64 \\
Bagel & 67.88 & 45.09 & 65.66 & 64.81 & 55.53 & 58.43 & 61.64 & 60.15 \\
\midrule
\multicolumn{9}{l}{\textit{CLEAR variants (Bagel backbone)}} \\
Text-only CoT & 63.62 & 48.30 & 70.33 & 64.18 & 56.93 & 53.98 & 62.82 & 60.02 \\
CLEAR-SFT & \underline{72.06} & 47.56 & \underline{70.33} & \underline{70.51} & \underline{57.67} & \underline{60.13} & \underline{65.65} & \underline{63.42} \\
CLEAR-RL & \textbf{72.52} & \textbf{51.97} & \textbf{71.33} & \textbf{72.25} & \textbf{60.67} & \textbf{61.05} & \textbf{67.07} & \textbf{65.26} \\
\bottomrule
\end{tabular}}
\end{table*}

\subsection{Interleaved GRPO}
\label{sec:grpo}

After SFT, the model can produce generate-then-answer trajectories, and the bridge provides a differentiable path from generation to reasoning.
The missing piece is a training signal that connects answer correctness to the generation process, so that the model learns to generate visual states that actually help it answer rather than simply approximate clean images under an MSE objective.
Interleaved GRPO fills this role by jointly optimizing text reasoning and visual generation under answer-correctness rewards.

\noindent\textbf{Background.}
GRPO~\cite{grpo} optimizes a language model by sampling a group of $G$ completions for each input, computing group-relative advantages from their rewards, and updating the policy with a clipped surrogate loss that increases the probability of higher-reward completions.
Flow-GRPO~\cite{flowgrpo} extends this idea to flow matching models by converting deterministic ODE sampling into an equivalent SDE~\cite{lipman2023flow, liu2023flow} to introduce the stochasticity that GRPO requires, and deriving per-step transition log-probabilities from the predicted velocity field so that the same clipped surrogate structure can be applied to denoising steps.

\noindent\textbf{Challenge of Joint Optimization.}
In our setting, each trajectory interleaves text tokens and a multi-step denoising process within a single autoregressive sequence, and we need to optimize both modalities under a shared reward.
Naively combining the two objectives would require maintaining the full computation graph across all $N$ denoising steps for each of the $G$ sampled trajectories, which is prohibitive in GPU memory since each denoising step involves a full forward pass through the model backbone.

\noindent\textbf{Trajectory Sampling and Training.}
We address this through two design choices that reduce the cost of image-side optimization to a tractable level.

For trajectory sampling, we generate $G$ complete interleaved sequences per input.
The text portion of each trajectory is sampled autoregressively as in standard GRPO.
For the denoising portion, each trajectory uses SDE-based sampling to generate a single denoising trajectory of $N$ steps, recording the state pair $(\mathbf{x}_t, \mathbf{x}_{t+\Delta t})$ at each step without retaining the computation graph.
The reward $R_i$ for each trajectory is computed from the final answer, and the group-relative advantage $\hat{A}_i$ is derived across the $G$ trajectories.

For the training forward pass, we randomly select one denoising step from the $N$ recorded states for each trajectory and inject the corresponding noisy latent $\mathbf{x}_t$ into the model input at its original position in the sequence.
The model then performs a single forward pass over the full interleaved sequence, simultaneously producing text logits at all text positions and the predicted velocity field $\mathbf{v}_\theta$ at the selected denoising position.
This reduces the image-side optimization from $N$ forward passes per trajectory to one, making the memory and compute cost comparable to standard text-only GRPO with only one additional token position per sequence.

From the text logits, we compute the standard GRPO loss:
\begin{equation}
\begin{split}
    \mathcal{L}_\text{GRPO} = -\mathbb{E}\Big[\min\Big(r_{i,t} \cdot \hat{A}_i, \\
    \text{clip}(r_{i,t}, 1{-}\epsilon, 1{+}\epsilon) \cdot \hat{A}_i\Big)\Big],
\end{split}
\end{equation}
where $r_{i,t} = \pi_\theta(o_{i,t} | q, o_{i,<t}) / \pi_{\theta_\text{old}}(o_{i,t} | q, o_{i,<t})$ is the per-token importance ratio.
From the predicted velocity field, we compute the transition log-probability under the SDE formulation and obtain the Flow-GRPO loss:
\begin{equation}
\begin{split}
    \mathcal{L}_\text{Flow-GRPO} = -\min\Big(r_\text{img} \cdot \hat{A}_i, \\
    \text{clip}(r_\text{img}, 1{-}\epsilon, 1{+}\epsilon) \cdot \hat{A}_i\Big),
\end{split}
\end{equation}
where $r_\text{img} = \exp(\log p_\theta(\mathbf{x}_{t+\Delta t} | \mathbf{x}_t) - \log p_{\theta_\text{old}}(\mathbf{x}_{t+\Delta t} | \mathbf{x}_t))$ is the transition probability ratio at the selected denoising step.
The final Interleaved GRPO loss combines both:
\begin{equation}
    \mathcal{L}_\text{Interleaved} = \mathcal{L}_\text{GRPO} + \lambda\,\mathcal{L}_\text{Flow-GRPO}.
\end{equation}
Because both losses are derived from the same forward pass and share hidden representations, gradients from the GRPO loss influence the image generation pathway through the bridge, and gradients from the Flow-GRPO loss influence textual reasoning through the shared attention mechanism.
Critically, both objectives use the same advantage $\hat{A}_i$ derived from a single reward, ensuring that text reasoning and visual generation are optimized toward the same goal.
By selecting only one denoising step per trajectory, the training forward pass adds minimal memory overhead beyond standard text-only GRPO while still coupling the two modalities within a shared computation graph.

\noindent\textbf{Reward Design.}
The reward combines three components.
The dominant term $R_\text{acc}$ measures final answer correctness, evaluated by an external language model following the LLM-as-judge paradigm~\cite{zheng2024judging} on a continuous scale.
$R_\text{fmt}$ encourages valid output structure by checking for properly formed analysis and answer blocks.
$R_\text{dec}$ evaluates the generation decision retrospectively: it assigns higher rewards when the model generated before answering correctly and penalizes cases where the model skipped generation and answered incorrectly; the remaining two cases (generated but answered incorrectly, or skipped generation and answered correctly) receive a neutral reward.
This encourages the model to invoke generation when it would help while not penalizing correct decisions to skip.
No reward component targets the perceptual quality of the generated visual state.
The overall reward is
\begin{equation}
    R = w_\text{acc}\,R_\text{acc} + w_\text{fmt}\,R_\text{fmt} + w_\text{dec}\,R_\text{dec}.
\end{equation}

\noindent\textbf{Adaptive Generation Strategy.}
The decision reward $R_\text{dec}$, combined with the natural mixture of generate-then-answer and direct-answer trajectories in the sampled completions, gives rise to an input-dependent generation policy.
During the analysis phase, the model implicitly evaluates whether generation would improve its answer and decides whether to emit the \texttt{<image\_restore>} token.
This is not a separate classifier or a manually designed threshold, but a behavior shaped by the reward signal within the Interleaved GRPO framework.
As we show in Section~\ref{sec:experiments}, the model learns to generate more frequently as degradation severity increases and to largely skip generation on clean inputs, achieving a favorable balance between robustness and efficiency.

\section{Experiments}
\label{sec:experiments}

\subsection{Implementation Details}

CLEAR is built on Bagel-7B~\cite{bagel} with a SigLIP~\cite{siglip} vision encoder and a Qwen2-based~\cite{qwen2} language model backbone.
Only the language model backbone is updated; the SigLIP encoder, VAE encoder, and VAE decoder remain frozen.
The SFT dataset contains 24k samples split evenly between direct-answer and generate-then-answer trajectories, constructed from LLaVA-OneVision~\cite{llavaonevision} as described in Section~\ref{sec:sft}.
We train SFT for 3 epochs with learning rate 2e-5, loss weights $\lambda_\text{MSE}{=}0.5$ and $\lambda_\text{KL}{=}0.1$, and ViT token drop probability 0.4.
For Interleaved GRPO, we use a separate 24k-sample set with group size $G{=}4$, learning rate 1e-6, $\epsilon{=}0.2$, image loss weight $\lambda{=}0.3$, and reward weights $w_\text{acc}{=}0.75$, $w_\text{fmt}{=}0.1$, $w_\text{dec}{=}0.15$.
Denoising uses 30 steps.
All experiments run on 8 NVIDIA A100 80GB GPUs.

We evaluate on MMD-Bench, which applies 16 corruption types at three severity levels (detailed in the supplementary material~\ref{app:mmdbench}) to six benchmarks: MMBench~\cite{mmbench}, MM-Vet~\cite{mmvet}, MMVP~\cite{mmvp}, CV-Bench~\cite{cvbench}, MMStar~\cite{mmstar}, and RealWorldQA, plus R-Bench-Dis~\cite{rbench} as an existing degraded-image benchmark.All evaluations are conducted using VLMEvalKit~\cite{duan2024vlmevalkit}.

\subsection{Main Results}

Table~\ref{tab:main} presents the main results under Hard degradation. We highlight three key observations.

\textbf{(1) Degradation vulnerability is universal.}
All models suffer substantial accuracy losses under degradation regardless of architecture and scale.
Even GPT-4.1-mini and Gemini-2.5-Flash show notable drops compared to their clean-image performance (Figure~\ref{fig:teaser}).
Among open-source unified models, Emu3, Janus-Pro, and Bagel all degrade significantly, confirming that \textit{existing generative pathways do not spontaneously contribute to robustness}.

\textbf{(2) Verbal reasoning cannot compensate for visual information loss.}
Text-only CoT provides no meaningful advantage over the base model (60.02 vs 60.15), with scattered gains on some benchmarks offset by regressions on others (e.g., MMBench 63.62 vs 67.88), indicating that \textit{fine-grained visual information destroyed by degradation cannot be recovered through language-level reasoning alone}.

\textbf{(3) Connecting generation to reasoning yields substantial gains.}
CLEAR-SFT improves the average by 3.27 points over Bagel with consistent gains across all benchmarks.
CLEAR-RL pushes this further to 65.26, the best result among all open-source models on all seven evaluation sets.
The gain from SFT to RL is most pronounced on MM-Vet (47.56 $\to$ 51.97) and MMStar (57.67 $\to$ 60.67), confirming the value of Interleaved GRPO for benchmarks requiring multi-cue reasoning.
Overall, \textit{CLEAR-RL improves Bagel by 5.11 points (8.5\% relative) within the same architecture without additional parameters or external modules}.

\textbf{(4) Comparison with external restoration.}
Although CLEAR addresses the internal capability disconnect within unified models rather than competing with external restoration pipelines, we provide a reference comparison on R-Bench~\cite{rbench}, an independently constructed degraded-image benchmark. A restoration model~\cite{chen2022simplebaselinesimagerestoration} followed by Bagel reaches 65.05, improving over the base Bagel (61.64) but still falling behind CLEAR-RL (67.07) by 2.02 points. The restoration model optimizes for pixel-level fidelity without coupling to the downstream reasoning task, whereas CLEAR shapes its generated states end-to-end through answer-correctness rewards, producing intermediate representations that better serve understanding.

\subsection{Robustness Analysis}

A natural question is whether CLEAR's gains under degradation simply reflect an overall quality improvement from fine-tuning or a genuine increase in robustness.
Table~\ref{tab:robustness} addresses this by comparing the performance drop from clean to hard inputs.

\begin{table}[t]
\centering
\caption{Robustness analysis. Clean and Hard scores are averaged over the six MMD-Bench benchmarks. Drop = Clean $-$ Hard.}
\label{tab:robustness}
\resizebox{0.6\linewidth}{!}{
\begin{tabular}{lccc}
\toprule
Method & Clean & Hard & Drop ($\downarrow$) \\
\midrule
Bagel & 66.86 & 59.57 & 7.29 \\
CLEAR-SFT & 69.34 & 63.04 & 6.30 \\
CLEAR-RL & 70.27 & 64.96 & 5.31 \\
\bottomrule
\end{tabular}}
\end{table}

Bagel loses 7.29 points (10.9\% relative) under hard degradation.
CLEAR-RL reduces this drop to 5.56 points (7.8\%), a 24\% reduction in the robustness gap.
The improvement on clean images reflects the benefit of the structured reasoning format shared by all fine-tuned variants, while the narrower degradation gap demonstrates the additional contribution of the generative pathway.
CLEAR's advantage over Bagel also widens as degradation severity increases, from +4.11 on clean to +5.39 on hard (full severity-level results in the supplement), directly confirming that the generative pathway provides increasing benefit when degradation is most severe.

\begin{figure*}[t]
    \centering
    \includegraphics[width=0.75\textwidth]{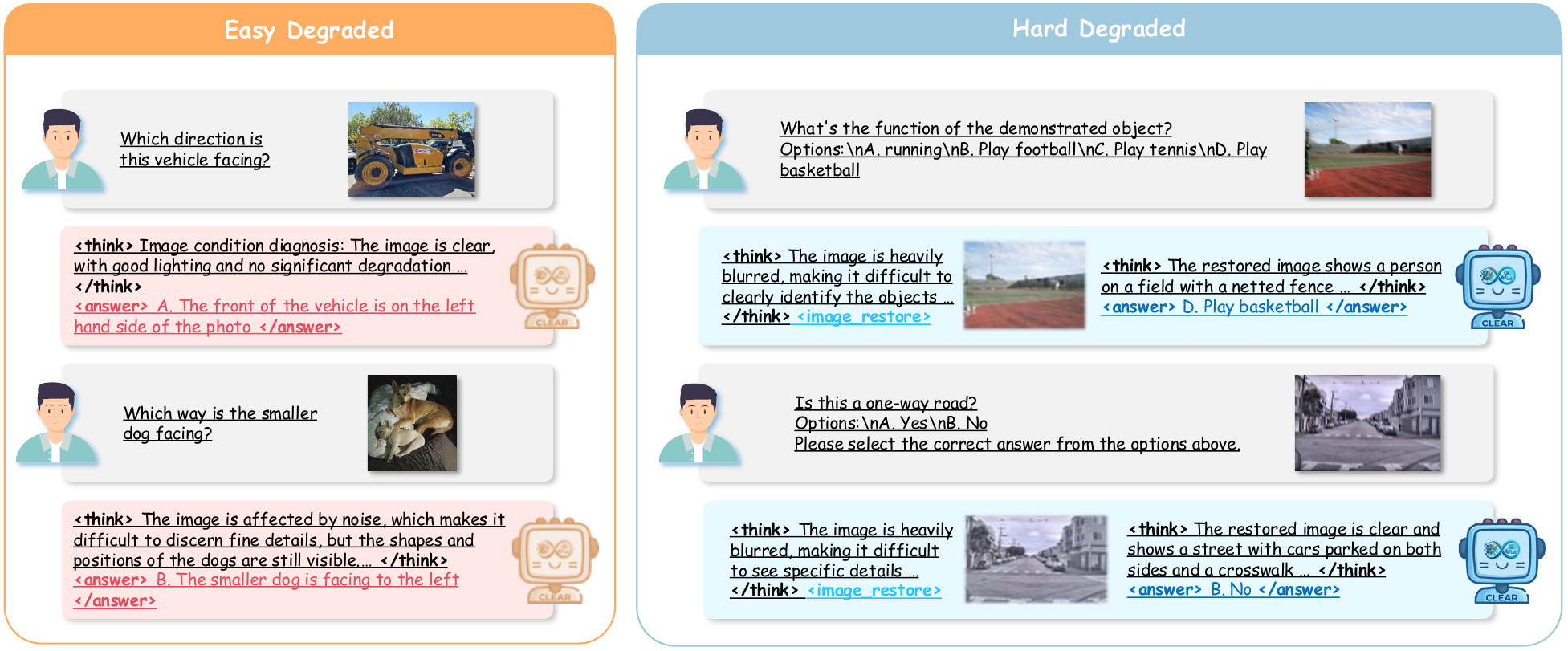}
        \caption{Qualitative examples of CLEAR's adaptive reasoning. Left: on a mildly noisy image, the model skips generation and answers directly. Right: on a severely blurred image, the model triggers generation to recover obscured details before answering.}
    \label{fig:qualitative}
\end{figure*}

\subsection{Ablation Studies}

Table~\ref{tab:ablation_steps} validates the necessity of each progressive step by systematically removing components.

\begin{table}[t]
\centering
\caption{Component ablation averaged over six MMD-Bench benchmarks. ``Dec-reenc'' replaces the bridge with the standard decode-reencode path during GRPO.}
\label{tab:ablation_steps}
\resizebox{0.6\linewidth}{!}{
\begin{tabular}{lcc}
\toprule
Configuration & Clean & Hard \\
\midrule
Bagel (base) & 66.86 & 59.57 \\
+ SFT & 69.34 & 63.04 \\
+ SFT + Dec-reenc + GRPO & 70.14 & 63.72 \\
+ SFT + Bridge (w/o GRPO) & 69.51 & 63.11 \\
+ SFT + Bridge + GRPO & \textbf{70.27} & \textbf{64.96} \\
\bottomrule
\end{tabular}}
\end{table}

\begin{table}[t]
\centering
\caption{No-reference perceptual quality and reasoning accuracy of intermediate visual states. BRISQUE and NIQE (lower is better) measure distortion; MUSIQ (higher is better) measures overall quality.}
\label{tab:perceptual}
\resizebox{0.8\linewidth}{!}{
\begin{tabular}{lcccc}
\toprule
State & BRISQUE$\downarrow$ & NIQE$\downarrow$ & MUSIQ$\uparrow$ & Hard AVG$\uparrow$ \\
\midrule
SFT state & 43.73 & 5.32 & 42.63 & 63.04 \\
RL state & \textbf{41.53} & \textbf{4.93} & \textbf{45.74} & \textbf{64.96} \\
\bottomrule
\end{tabular}}
\end{table}

\begin{figure}[t]
    \centering
    \includegraphics[width=\linewidth]{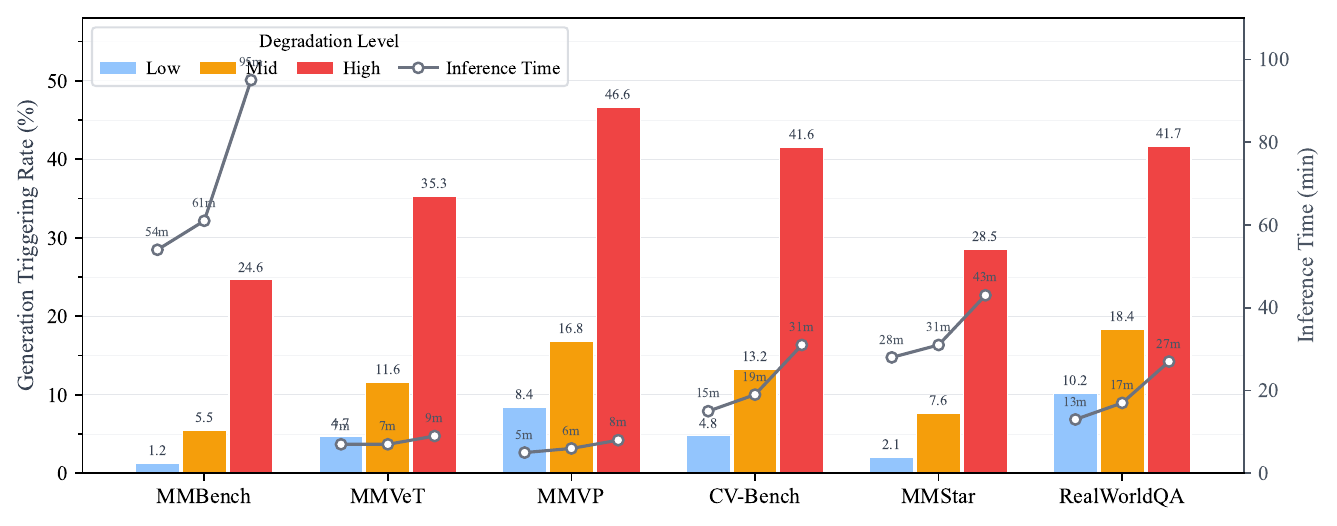}
\caption{Generation triggering rate (bars, left axis) and total inference time (line, right axis) across degradation severity levels for each benchmark.}
    \label{fig:trigger_and_cost}
\end{figure}

Applying GRPO directly to the base Bagel model without SFT is not feasible, because Bagel has never been trained to produce generate-then-answer trajectories.
Without the behavioral pattern established by SFT, the model does not emit the \texttt{<image\_restore>} token or structure its output in the interleaved format that GRPO requires, leaving the reinforcement learning process without valid trajectories to optimize.

SFT alone yields a 3.47\% gain on hard inputs, demonstrating that the generate-then-answer pattern is valuable even without joint optimization.
Replacing the bridge with decode-reencode during GRPO limits the gain to 63.72, because the frozen decoder and encoder block answer-level credit from reaching the generation process.
The bridge without GRPO performs comparably to SFT alone (63.11 vs 63.04), confirming that its value lies in enabling joint optimization rather than providing a better inference-time representation.
The full pipeline achieves the best result on both clean and hard inputs, with each component building on the previous one.

Beyond validating each component, we further examine whether the generated VAE latent provides visual information that the degraded input alone cannot supply. We keep the full generate-then-answer trajectory intact but replace the generated latent at the bridge with the degraded image's own VAE latent, preserving the same token format, count, and position in the reasoning context. Under this substitution the hard-degradation average drops from 64.96 to 62.06. Since the reasoning structure remains identical and only the latent content differs, this gap confirms that the denoising process recovers visual structure absent from the degraded input and that the model's subsequent reasoning actively relies on this recovered information.

\subsection{Analysis}

\textbf{Adaptive Generation Behavior and Inference Overhead.}
Figure~\ref{fig:trigger_and_cost} shows the generation triggering rate and total inference time across degradation levels. The average triggering rate rises monotonically from 5.2\% at low to 12.2\% at mid and 36.4\% at high, with MMVP and RealWorldQA reaching the highest rates (46.6\% and 41.7\%) due to their reliance on fine-grained visual detail. Inference time closely tracks the triggering rate: at low degradation, evaluation time remains near the base model, while under high degradation the additional denoising cost raises time in proportion to the fraction of samples that trigger generation. The overhead is thus determined by the adaptive policy rather than any fixed per-input cost, confirming that CLEAR concentrates computation on inputs where generation yields the largest accuracy benefit.

\noindent\textbf{Intermediate Visual States.}
A central claim of this work is that pixel-level reconstruction supervision constrains rather than helps the generation process.
To test this, we evaluate no-reference perceptual quality metrics on samples that triggered generation under hard degradation (Table~\ref{tab:perceptual}).
During SFT, the MSE loss encourages generated states to approximate the clean image, but its well-known regression-to-mean tendency produces perceptually smooth outputs that score poorly on sharpness and texture metrics.
After Interleaved GRPO, pixel-level supervision is removed entirely and generation is driven solely by answer-correctness rewards.
Despite receiving no perceptual quality signal, RL states consistently outperform SFT states across all three metrics, because the visual properties that help reasoning, sharp edges for reading text, clear textures for identifying objects, well-defined structure for spatial reasoning, are precisely those that no-reference metrics also value.
A pure task-driven reward therefore simultaneously improves both reasoning accuracy and perceptual quality, confirming that the two objectives are naturally aligned and that explicit reconstruction supervision acts as a constraint rather than a requirement.

\noindent\textbf{Qualitative Examples.}
Figure~\ref{fig:qualitative} illustrates CLEAR's reasoning in two contrasting scenarios.
On a mildly noisy image, the model judges that the available visual information is sufficient, skips generation entirely, and answers directly.
On a severely blurred image, the first analysis phase identifies that critical visual details are unreadable, triggers generation, and the post-generation phase extracts recovered information for a correct answer.

\section{Conclusion}

We identified a functional disconnect in unified multimodal models where generation and understanding coexist but remain isolated under degraded inputs, and proposed CLEAR to bridge this gap.
Through supervised fine-tuning that establishes the generate-then-answer reasoning pattern, a Latent Representation Bridge that opens a direct optimization route from generation to reasoning, and Interleaved GRPO that jointly optimizes both capabilities under answer-correctness rewards, CLEAR enables unified models to leverage their own generative capacity for robust visual understanding.
Experiments on MMD-Bench show that CLEAR substantially improves degraded-image performance while preserving clean-image accuracy, with the model learning to invoke generation selectively based on input quality.
Our analysis further reveals that removing pixel-level reconstruction supervision and relying solely on answer-correctness rewards leads to intermediate visual states with higher perceptual quality, not lower, confirming that task-driven optimization and visual clarity are naturally aligned and that explicit reconstruction targets act as a constraint rather than a requirement.

{
    \small
    \bibliographystyle{ieeenat_fullname}
    \bibliography{main}

@String(CVPR= {IEEE Conf. Comput. Vis. Pattern Recog.})

@String(ECCV= {Eur. Conf. Comput. Vis.})

@String(ICLR = {Int. Conf. Learn. Represent.})

@String(CVPR  = {CVPR})

@String(ECCV  = {ECCV})

@String(ICLR  = {ICLR})

@inproceedings{llava,
  title={Visual Instruction Tuning},
  author={Liu, Haotian and Li, Chunyuan and Wu, Qingyang and Lee, Yong Jae},
  booktitle={NeurIPS},
  year={2023}
}

@article{qwen2vl,
  title={Qwen2-VL: Enhancing Vision-Language Model's Perception of the World at Any Resolution},
  author={Wang, Peng and Bai, Shuai and Tan, Sinan and Wang, Shijie and Fan, Zhihao and Bai, Jinze and Chen, Keqin and Liu, Xuejing and Wang, Jialin and Ge, Wenbin and Fan, Yang and Dang, Kai and Du, Mengfei and Ren, Xuancheng and Men, Rui and Liu, Dayiheng and Zhou, Chang and Zhou, Jingren and Lin, Junyang},
  journal={arXiv preprint arXiv:2409.12191},
  year={2024}
}

@inproceedings{internvl,
  title={InternVL: Scaling up Vision Foundation Models and Aligning for Generic Visual-Linguistic Tasks},
  author={Chen, Zhe and Wu, Jiannan and Wang, Wenhai and Su, Weijie and Chen, Guo and Xing, Sen and Zhong, Muyan and Zhang, Qinglong and Zhu, Xizhou and Lu, Lewei and Li, Bin and Luo, Ping and Lu, Tong and Qiao, Yu and Dai, Jifeng},
  booktitle={Proceedings of the IEEE/CVF Conference on Computer Vision and Pattern Recognition},
  pages={24185--24198},
  year={2024}
}

@inproceedings{clip,
  title={Learning Transferable Visual Models From Natural Language Supervision},
  author={Radford, Alec and Kim, Jong Wook and Hallacy, Chris and Ramesh, Aditya and Goh, Gabriel and Agarwal, Sandhini and Sastry, Girish and Askell, Amanda and Mishkin, Pamela and Clark, Jack and Krueger, Gretchen and Sutskever, Ilya},
  booktitle={International Conference on Machine Learning},
  pages={8748--8763},
  year={2021},
  organization={PMLR}
}

@inproceedings{siglip,
  title={Sigmoid Loss for Language Image Pre-Training},
  author={Zhai, Xiaohua and Mustafa, Basil and Kolesnikov, Alexander and Beyer, Lucas},
  booktitle={Proceedings of the IEEE/CVF International Conference on Computer Vision},
  year={2023}
}

@article{bagel,
  title={Emerging Properties in Unified Multimodal Pretraining},
  author={Li, Kunchang and others},
  journal={arXiv preprint arXiv:2505.14683},
  year={2025}
}

@article{janus,
  title={Janus: Decoupling Visual Encoding for Unified Multimodal Understanding and Generation},
  author={Wu, Chengyue and Chen, Xiaokang and Wu, Zhiyu and Ma, Yiyang and Liu, Xingchao and Pan, Zizheng and Liu, Wen and Xie, Zhenda and Yu, Xingkai and Ruan, Chong and Luo, Ping},
  journal={arXiv preprint arXiv:2410.13848},
  year={2024}
}

@article{emu3,
  title={Emu3: Next-Token Prediction is All You Need},
  author={Wang, Xinlong and Zhang, Xiaosong and Luo, Zhengxiong and Sun, Quan and Cui, Yufeng and Wang, Jinsheng and Zhang, Fan and Wang, Yueze and Li, Zhen and Yu, Qiying and others},
  journal={arXiv preprint arXiv:2409.18869},
  year={2024}
}

@article{chameleon,
  title={Chameleon: Mixed-Modal Early-Fusion Foundation Models},
  author={{Chameleon Team}},
  journal={arXiv preprint arXiv:2405.09818},
  year={2024}
}

@article{imagenetc,
  title={Benchmarking Neural Network Robustness to Common Corruptions and Perturbations},
  author={Hendrycks, Dan and Dietterich, Thomas},
  journal={Proceedings of the International Conference on Learning Representations},
  year={2019}
}

@article{rbench,
  title={R-Bench: Are your Large Multimodal Model Robust to Real-world Corruptions?},
  author={Li, Chunyi and Zhang, Jianbo and Zhang, Zicheng and Wu, Haoning and Tian, Yuan and Sun, Wei and Lu, Guo and Liu, Xiaohong and Min, Xiongkuo and Lin, Weisi and Zhai, Guangtao},
  journal={arXiv preprint arXiv:2410.05474},
  year={2024}
}

@article{mmebench,
  title={MME: A Comprehensive Evaluation Benchmark for Multimodal Large Language Models},
  author={Fu, Chaoyou and Chen, Peixian and Shen, Yunhang and Qin, Yulei and Zhang, Mengdan and Lin, Xu and Yang, Jinrui and Zheng, Xiawu and Li, Ke and Sun, Xing and Wu, Yunsheng and Ji, Rongrong},
  journal={arXiv preprint arXiv:2306.13394},
  year={2023}
}

@article{zhao2024evaluating,
  title={Evaluating the Robustness of Multimodal Large Language Models Against Image Corruptions},
  author={Zhao, Changqian and others},
  journal={arXiv preprint},
  year={2024},
}

@inproceedings{restormer,
  title={Restormer: Efficient Transformer for High-Resolution Image Restoration},
  author={Zamir, Syed Waqas and Arora, Aditya and Khan, Salman and Hayat, Munawar and Khan, Fahad Shahbaz and Yang, Ming-Hsuan},
  booktitle={CVPR},
  year={2022}
}

@inproceedings{nafnet,
  title={Simple Baselines for Image Restoration},
  author={Chen, Liangyu and Chu, Xiaojie and Zhang, Xiangyu and Sun, Jian},
  booktitle={European Conference on Computer Vision (ECCV)},
  year={2022}
}

@article{vila-u,
  title={VILA-U: a Unified Foundation Model Integrating Visual Understanding and Generation},
  author={Wu, Yecheng and Zhang, Zhuoyang and Chen, Junyu and Tang, Haotian and Li, Dacheng and Fang, Yunhao and Zhu, Ligeng and Xie, Enze and Yin, Hongxu and Yi, Li and Han, Song and Lu, Yao},
  journal={arXiv preprint arXiv:2409.04429},
  year={2024}
}

@article{show-o,
  title={Show-o: One Single Transformer to Unify Multimodal Understanding and Generation},
  author={Xie, Jinheng and Mao, Weijia and Bai, Zechen and Zhang, David Junhao and Wang, Weihao and Lin, Kevin Qinghong and Gu, Yuchao and Chen, Zhijie and Yang, Zhenheng and Shou, Mike Zheng},
  journal={arXiv preprint arXiv:2408.12528},
  year={2024}
}

@article{transfusion,
  title={Transfusion: Predict the Next Token and Diffuse Images with One Multi-Modal Model},
  author={Zhou, Chunting and Yu, Lili and Babu, Arun and Tirumala, Kushal and Yasunaga, Michihiro and Shamis, Leonid and Kahn, Jacob and Ma, Xuezhe and Zettlemoyer, Luke and Levy, Omer},
  journal={arXiv preprint arXiv:2408.11039},
  year={2024}
}

@article{grpo,
  title={DeepSeekMath: Pushing the Limits of Mathematical Reasoning in Open Language Models},
  author={Shao, Zhihong and Wang, Peiyi and Zhu, Qihao and Xu, Runxin and Song, Junxiao and Zhang, Mingchuan and Li, Y.K. and Wu, Y. and Guo, Daya},
  journal={arXiv preprint arXiv:2402.03300},
  year={2024}
}

@article{deepseekr1,
  title={DeepSeek-R1: Incentivizing Reasoning Capability in LLMs via Reinforcement Learning},
  author={{DeepSeek-AI}},
  journal={arXiv preprint arXiv:2501.12948},
  year={2025}
}

@article{flowgrpo,
  title={Improving Generation Quality of Flow-based Multimodal Models via GRPO},
  author={},
  journal={arXiv preprint},
  year={2025},
}

@article{ddpo,
  title={Training Diffusion Models with Reinforcement Learning},
  author={Black, Kevin and Janner, Michael and Du, Yilun and Kostrikov, Ilya and Levine, Sergey},
  journal={arXiv preprint arXiv:2305.13301},
  year={2023}
}

@article{gpt4o,
  title={GPT-4o System Card},
  author={OpenAI},
  journal={arXiv preprint arXiv:2410.21276},
  year={2024}
}

@article{kingma2013auto,
  title={Auto-Encoding Variational Bayes},
  author={Kingma, Diederik P and Welling, Max},
  journal={arXiv preprint arXiv:1312.6114},
  year={2013}
}

@inproceedings{rombach2022high,
  title={High-Resolution Image Synthesis with Latent Diffusion Models},
  author={Rombach, Robin and Blattmann, Andreas and Lorenz, Dominik and Esser, Patrick and Ommer, Bj{\"o}rn},
  booktitle={CVPR},
  year={2022}
}

@inproceedings{esser2021taming,
  title={Taming Transformers for High-Resolution Image Synthesis},
  author={Esser, Patrick and Rombach, Robin and Ommer, Bj{\"o}rn},
  booktitle={CVPR},
  year={2021}
}

@inproceedings{hendrycks2020augmix,
  title={AugMix: A Simple Data Processing Method to Improve Robustness and Uncertainty},
  author={Hendrycks, Dan and Mu, Norman and Cubuk, Ekin D and Zoph, Barret and Gilmer, Justin and Lakshminarayanan, Balaji},
  booktitle={ICLR},
  year={2020}
}

@inproceedings{mintun2021interaction,
  title={On Interaction Between Augmentations and Corruptions in Natural Corruption Robustness},
  author={Mintun, Eric and Kirillov, Alexander and Xie, Saining},
  booktitle={NeurIPS},
  year={2021}
}

@inproceedings{swinir,
  title={SwinIR: Image Restoration Using Swin Transformer},
  author={Liang, Jingyun and Cao, Jiezhang and Sun, Guolei and Zhang, Kai and Van Gool, Luc and Timofte, Radu},
  booktitle={ICCVW},
  year={2021}
}

@article{unifiedio2,
  title={Unified-IO 2: Scaling Autoregressive Multimodal Models with Vision, Language, Audio, and Action},
  author={Lu, Jiasen and Clark, Christopher and Lee, Sangho and Zhang, Zichen and Khosla, Savya and Marten, Ryan and Hoiem, Derek and Kembhavi, Aniruddha},
  journal={CVPR},
  year={2024}
}

@article{diffusiondpo,
  title={Diffusion Model Alignment Using Direct Preference Optimization},
  author={Wallace, Bram and Dang, Meiqi and Rafailov, Rafael and Zhou, Linqi and Lou, Aaron and Purber, Senthil and Ermon, Stefano and Xiong, Caiming and Joty, Shafiq and Naik, Nikhil},
  booktitle={CVPR},
  year={2024}
}

@article{llavaonevision,
  title={LLaVA-OneVision: Easy Visual Task Transfer},
  author={Li, Bo and Zhang, Yuanhan and Guo, Dong and Zhang, Renrui and Li, Feng and Zhang, Hao and Zhang, Kaichen and Li, Yanwei and Liu, Ziwei and Li, Chunyuan},
  journal={arXiv preprint arXiv:2408.03326},
  year={2024}
}

@article{visionr1,
  title={Vision-R1: Incentivizing Reasoning Capability in Multimodal Large Language Models},
  author={Huang, Wenyi and Feng, Enfang and Gao, Yufei and others},
  journal={arXiv preprint arXiv:2503.06749},
  year={2025}
}

@article{vlmr1,
  title={VLM-R1: A Stable and Generalizable R1-style Large Vision-Language Model},
  author={Shen, Haozhan and Zhang, Zilun and Zhao, Qian and Zhang, Ruochen and others},
  journal={arXiv preprint arXiv:2504.07615},
  year={2025}
}

@article{r1v,
  title={R1-V: Reinforcing Super Generalization Ability in Vision Language Models with Less Than \$3},
  author={Chen, Liang and Bai, Qiguang and Xu, Kanzhi and Li, Jiahao and others},
  journal={arXiv preprint arXiv:2503.01785},
  year={2025}
}

@misc{gpt41,
  title={GPT-4.1},
  author={OpenAI},
  howpublished={\url{https://openai.com/index/gpt-4-1/}},
  year={2025}
}

@inproceedings{mathieu2016deep,
  title={Deep Multi-Scale Video Prediction Beyond Mean Square Error},
  author={Mathieu, Michael and Couprie, Camille and LeCun, Yann},
  booktitle={ICLR},
  year={2016}
}

@inproceedings{ledig2017photo,
  title={Photo-Realistic Single Image Super-Resolution Using a Generative Adversarial Network},
  author={Ledig, Christian and Theis, Lucas and Husz{\'a}r, Ferenc and Caballero, Jose and Cunningham, Andrew and Acosta, Alejandro and Aitken, Andrew and Tejani, Alykhan and Totz, Johannes and Wang, Zehan and Shi, Wenzhe},
  booktitle={CVPR},
  year={2017}
}

@inproceedings{lipman2023flow,
  title={Flow Matching for Generative Modeling},
  author={Lipman, Yaron and Chen, Ricky T. Q. and Ben-Hamu, Heli and Nickel, Maximilian and Le, Matt},
  booktitle={ICLR},
  year={2023}
}

@inproceedings{liu2023flow,
  title={Flow Straight and Fast: Learning to Generate and Transfer Data with Rectified Flow},
  author={Liu, Xingchao and Gong, Chengyue and Liu, Qiang},
  booktitle={ICLR},
  year={2023}
}

@inproceedings{zheng2024judging,
  title={Judging LLM-as-a-Judge with MT-Bench and Chatbot Arena},
  author={Zheng, Lianmin and Chiang, Wei-Lin and Sheng, Ying and Zhuang, Siyuan and Wu, Zhanghao and Zhuang, Yonghao and Lin, Zi and Li, Zhuohan and Li, Dacheng and Xing, Eric P. and Zhang, Hao and Gonzalez, Joseph E. and Stoica, Ion},
  booktitle={NeurIPS},
  year={2024}
}

@article{qwen2,
  title={Qwen2 Technical Report},
  author={Yang, An and Yang, Baosong and Hui, Binyuan and Zheng, Bo and Yu, Bowen and Zhou, Chang and others},
  journal={arXiv preprint arXiv:2407.10671},
  year={2024}
}

@misc{realworldqa,
  title={RealWorldQA},
  author={xAI},
  howpublished={\url{https://huggingface.co/datasets/xai-org/RealWorldQA}},
  year={2024}
}

@article{mmbench,
  title={MMBench: Is Your Multi-modal Model an All-around Player?},
  author={Liu, Yuan and Duan, Haodong and Zhang, Yuanhan and Li, Bo and Zhang, Songyang and Zhao, Wangbo and Yuan, Yike and Wang, Jiaqi and He, Conghui and Liu, Ziwei and Chen, Kai and Lin, Dahua},
  journal={arXiv preprint arXiv:2307.06281},
  year={2023}
}

@article{mmvet,
  title={MM-Vet: Evaluating Large Multimodal Models for Integrated Capabilities},
  author={Yu, Weihao and Yang, Zhengyuan and Li, Linjie and Wang, Jianfeng and Lin, Kevin and Liu, Zicheng and Wang, Xinchao and Wang, Lijuan},
  journal={arXiv preprint arXiv:2308.02490},
  year={2023}
}

@article{mmvp,
  title={Eyes Wide Shut? Exploring the Visual Shortcomings of Multimodal LLMs},
  author={Tong, Shengbang and Liu, Zhuang and Zhai, Yuexiang and Ma, Yi and LeCun, Yann and Xie, Saining},
  journal={CVPR},
  year={2024}
}

@article{cvbench,
  title={Cambrian-1: A Fully Open, Vision-Centric Exploration of Multimodal LLMs},
  author={Tong, Shengbang and Brown, Ellis and Wu, Penghao and Woo, Sanghyun and Middepogu, Manoj and Akula, Sai Charitha and Yang, Jihan and Yang, Shusheng and Iyer, Adithya and Pan, Xichen and others},
  journal={arXiv preprint arXiv:2406.16860},
  year={2024}
}

@article{mmstar,
  title={Are We on the Right Way for Evaluating Large Vision-Language Models?},
  author={Chen, Lin and Li, Jinsong and Dong, Xiaoyi and Zhang, Pan and Zang, Yuhang and Chen, Zehui and Duan, Haodong and Wang, Jiaqi and Qiao, Yu and Lin, Dahua and Zhu, Feng},
  journal={arXiv preprint arXiv:2403.20330},
  year={2024}
}

@article{zhang2025cooper,
  title={COOPER: A Unified Model for Cooperative Perception and Reasoning in Spatial Intelligence},
  author={Zhang, Zefeng and Hao, Xiangzhao and Tang, Hengzhu and Zhang, Zhenyu and Sheng, Jiawei and Li, Xiaodong and Li, Zhenyang and Gao, Li and Shi, Daiting and Yin, Dawei and others},
  journal={arXiv preprint arXiv:2512.04563},
  year={2025}
}

@article{zhang2023tale,
  title={A tale of two features: Stable diffusion complements dino for zero-shot semantic correspondence},
  author={Zhang, Junyi and Herrmann, Charles and Hur, Junhwa and Polania Cabrera, Luisa and Jampani, Varun and Sun, Deqing and Yang, Ming-Hsuan},
  journal={Advances in Neural Information Processing Systems},
  volume={36},
  pages={45533--45547},
  year={2023}
}

@inproceedings{duan2024vlmevalkit,
  title={Vlmevalkit: An open-source toolkit for evaluating large multi-modality models},
  author={Duan, Haodong and Yang, Junming and Qiao, Yuxuan and Fang, Xinyu and Chen, Lin and Liu, Yuan and Dong, Xiaoyi and Zang, Yuhang and Zhang, Pan and Wang, Jiaqi and others},
  booktitle={Proceedings of the 32nd ACM International Conference on Multimedia},
  pages={11198--11201},
  year={2024}
}

@misc{chen2022simplebaselinesimagerestoration,
      title={Simple Baselines for Image Restoration}, 
      author={Liangyu Chen and Xiaojie Chu and Xiangyu Zhang and Jian Sun},
      year={2022},
      eprint={2204.04676},
      archivePrefix={arXiv},
      primaryClass={cs.CV},
      url={https://arxiv.org/abs/2204.04676}, 
}
}


\clearpage
\section{Appendix}
\appendix

This supplementary material is organized as follows.
Appendix~\ref{app:grpo_detail} provides detailed derivations of the GRPO and Flow-GRPO objectives that underlie Interleaved GRPO.
Appendix~\ref{app:mmdbench} describes the construction of MMD-Bench, including the 16 corruption types and six base benchmarks.
Appendix~\ref{app:data} details the training data construction pipeline and reasoning trace generation process.
Appendix~\ref{app:prompt} presents the system prompt shared across training and inference.
Appendix~\ref{app:severity} reports full severity-level results.
Appendix~\ref{app:percorruption} provides per-corruption analysis.
Appendix~\ref{app:latency} analyzes inference latency.
Appendix~\ref{app:hyperparameter} examines hyperparameter sensitivity.
Appendix~\ref{app:reward} gives additional reward design details.
Appendix~\ref{app:qualitative} presents reasoning trace examples and additional qualitative results.

\section{GRPO and Flow-GRPO Details}
\label{app:grpo_detail}

This section provides the full computation process of GRPO and Flow-GRPO, which are combined into Interleaved GRPO in Section~\ref{sec:grpo} of the main text.

\subsection{GRPO}

Group Relative Policy Optimization~\cite{grpo} eliminates the need for a separate value network by estimating advantages from a group of sampled completions.
For each input query $q$, the model samples a group of $G$ completions $\{o_1, o_2, \ldots, o_G\}$ from the current policy $\pi_{\theta_\text{old}}$.
Each completion $o_i$ is scored by a reward function to obtain $R_i$.

The group-relative advantage for the $i$-th completion is computed by normalizing rewards within the group:
\begin{equation}
    \hat{A}_i = \frac{R_i - \text{mean}(R_1, R_2, \ldots, R_G)}{\text{std}(R_1, R_2, \ldots, R_G)}.
\end{equation}
This relative normalization ensures that the advantage reflects how good a completion is compared to its peers from the same input, rather than in absolute terms.

The policy is then updated by maximizing the clipped surrogate objective.
For each token $t$ in completion $o_i$, the per-token importance ratio is:
\begin{equation}
    r_{i,t} = \frac{\pi_\theta(o_{i,t} \mid q, o_{i,<t})}{\pi_{\theta_\text{old}}(o_{i,t} \mid q, o_{i,<t})}.
\end{equation}
The GRPO objective is:
\begin{equation}
\begin{split}
    \mathcal{J}_\text{GRPO}(\theta) = &\frac{1}{G}\sum_{i=1}^{G}\frac{1}{|o_i|}\sum_{t=1}^{|o_i|} \min\Big(r_{i,t} \cdot \hat{A}_i, \\
    &\text{clip}(r_{i,t}, 1{-}\epsilon, 1{+}\epsilon) \cdot \hat{A}_i\Big) - \beta\, D_\text{KL}[\pi_\theta \| \pi_\text{ref}],
\end{split}
\end{equation}
where $\epsilon$ is the clipping range that prevents excessively large policy updates, and the KL divergence term with coefficient $\beta$ regularizes the updated policy to stay close to a reference policy $\pi_\text{ref}$, preventing reward hacking.
Letting $\rho_{i,t} = \pi_\text{ref}(o_{i,t} \mid q, o_{i,<t}) / \pi_\theta(o_{i,t} \mid q, o_{i,<t})$, the KL divergence is estimated per token as:
\begin{equation}
    D_\text{KL}[\pi_\theta \| \pi_\text{ref}] \approx \rho_{i,t} - \log \rho_{i,t} - 1.
\end{equation}

The key advantage of GRPO over PPO is that no critic network is needed.
The group-relative advantage estimation replaces the learned value baseline with a simple statistical normalization over the sampled group, significantly reducing memory consumption and implementation complexity.

\subsection{Flow-GRPO}

Flow-GRPO~\cite{flowgrpo} extends GRPO to flow matching models, which generate images through a learned velocity field that transports samples from noise to data along a continuous-time trajectory.

\textbf{Flow Matching Background.}
In rectified flow~\cite{lipman2023flow, liu2023flow}, a velocity field $\mathbf{v}_\theta(\mathbf{x}_t, t)$ is learned to transport a noise sample $\mathbf{x}_1 \sim \mathcal{N}(\mathbf{0}, \mathbf{I})$ to a data sample $\mathbf{x}_0$ along a straight path.
The sampling process follows the ODE:
\begin{equation}
    d\mathbf{x}_t = \mathbf{v}_\theta(\mathbf{x}_t, t)\, dt,
\end{equation}
where $t$ decreases from 1 (pure noise) to 0 (clean image).
This deterministic process generates images by discretizing the ODE into $N$ steps.

\textbf{ODE-to-SDE Conversion.}
GRPO requires stochastic sampling to generate diverse trajectories for advantage estimation.
Since the flow ODE is deterministic, Flow-GRPO converts it into an equivalent SDE that preserves the same marginal distribution $p_t(\mathbf{x})$ at all timesteps.
Using the Fokker-Planck equation to match marginal densities, the equivalent reverse-time SDE is:
\begin{equation}
    d\mathbf{x}_t = \left[\mathbf{v}_\theta(\mathbf{x}_t, t) + \frac{\sigma_t^2}{2}\nabla \log p_t(\mathbf{x}_t)\right] dt + \sigma_t\, d\mathbf{w},
\end{equation}
where $\sigma_t$ is a noise schedule that controls the level of stochasticity and $d\mathbf{w}$ is a Wiener process.
The marginal score $\nabla \log p_t(\mathbf{x})$ is related to the velocity field by:
\begin{equation}
    \nabla \log p_t(\mathbf{x}) = -\frac{\mathbf{x}_t + (1-t)\mathbf{v}_\theta(\mathbf{x}_t, t)}{t}.
\end{equation}
Substituting this into the SDE and applying Euler-Maruyama discretization yields the update rule:
\begin{equation}
\begin{split}
    \mathbf{x}_{t+\Delta t} = \mathbf{x}_t + \Big[\mathbf{v}_\theta + \frac{\sigma^2}{2} \cdot \frac{\mathbf{x}_t + (1-t)\mathbf{v}_\theta}{t}\Big]\Delta t \\
    + \sigma\sqrt{\Delta t}\;\boldsymbol{\epsilon}, \quad \boldsymbol{\epsilon} \sim \mathcal{N}(\mathbf{0}, \mathbf{I}).
\end{split}
\end{equation}

\textbf{Transition Log-Probability.}
The SDE update defines a Gaussian transition distribution.
Letting $\mathbf{s}_\theta = -(\mathbf{x}_t + (1-t)\mathbf{v}_\theta) / t$ denote the score estimate, the predicted mean of the next state is:
\begin{equation}
    \boldsymbol{\mu}_\theta = \mathbf{x}_t + \left(\mathbf{v}_\theta - \tfrac{\sigma^2}{2}\mathbf{s}_\theta\right)\Delta t,
\end{equation}
and the transition log-probability is:
\begin{equation}
    \log p_\theta(\mathbf{x}_{t+\Delta t} \mid \mathbf{x}_t) = -\frac{1}{2\sigma^2 \Delta t}\left\|\mathbf{x}_{t+\Delta t} - \boldsymbol{\mu}_\theta\right\|^2 + C,
\end{equation}
where $C$ is a constant independent of $\theta$ that cancels in the importance ratio.

\textbf{Policy Update.}
Analogous to GRPO, Flow-GRPO samples $G$ denoising trajectories for each input, computes the reward for each, and derives the group-relative advantage $\hat{A}_i$.
The importance ratio at a denoising step is:
\begin{equation}
    r_\text{img} = \exp\left(\log p_\theta(\mathbf{x}_{t+\Delta t} \mid \mathbf{x}_t) - \log p_{\theta_\text{old}}(\mathbf{x}_{t+\Delta t} \mid \mathbf{x}_t)\right),
\end{equation}
and the Flow-GRPO objective follows the same clipped surrogate structure:
\begin{equation}
    \mathcal{L}_\text{Flow-GRPO} = -\min\left(r_\text{img} \cdot \hat{A}_i, \; \text{clip}(r_\text{img}, 1{-}\epsilon, 1{+}\epsilon) \cdot \hat{A}_i\right).
\end{equation}

\subsection{From Separate to Interleaved}

Standard GRPO operates on text-only sequences, while Flow-GRPO operates on image-only denoising trajectories.
In our setting, each trajectory contains both text tokens and a denoising process interleaved within a single autoregressive sequence.
The challenge is that these two objectives operate on fundamentally different token types (discrete text tokens vs. continuous latent states) yet must share the same reward signal.

Interleaved GRPO addresses this by making three design choices.
First, both objectives share the same group-relative advantage $\hat{A}_i$ computed from a single reward that evaluates the final answer, ensuring that text reasoning and visual generation are optimized toward the same goal.
Second, only one denoising step per trajectory is selected for optimization during training, reducing the memory cost from $N$ forward passes to one while still coupling the two modalities through shared hidden representations.
Third, the Latent Representation Bridge provides the differentiable connection that allows gradients from the text-side GRPO loss to reach the generation parameters and gradients from the Flow-GRPO loss to influence text reasoning.
The full formulation is presented in Section~\ref{sec:grpo} of the main text.

\section{MMD-Bench Details}
\label{app:mmdbench}

\subsection{Motivation and Comparison with R-Bench}

R-Bench~\cite{rbench} is the most closely related existing benchmark for evaluating multimodal models under image degradation.
While R-Bench provides a valuable testbed, it has several limitations that motivate the construction of MMD-Bench.

First, R-Bench uses a fixed set of pre-degraded images without providing clean counterparts or systematic severity control.
This makes it difficult to measure the performance drop caused by degradation or to analyze how model robustness changes as severity increases.
MMD-Bench addresses this by applying degradations to existing benchmarks whose clean-image performance is already well established, enabling direct computation of the clean-to-degraded gap.

Second, R-Bench does not organize its degradation types into structured categories that reflect real-world degradation sources, making it harder to diagnose which types of degradation a model is most vulnerable to.
MMD-Bench groups 16 corruption types into four categories (capture, transmission, environmental, and post-processing), enabling systematic analysis at both the category level and the individual corruption level.

Third, R-Bench evaluates a single combined score without separating the contributions of different visual capabilities.
By building on six established benchmarks that each target different aspects of multimodal understanding, MMD-Bench enables fine-grained diagnosis of which capabilities are most affected by degradation.

Table~\ref{tab:bench_comparison} summarizes the key differences.

\begin{table}[h]
\centering
\caption{Comparison between R-Bench and MMD-Bench.}
\label{tab:bench_comparison}
\resizebox{\linewidth}{!}{
\begin{tabular}{lcc}
\toprule
Property & R-Bench & MMD-Bench \\
\midrule
Clean reference available & No & Yes \\
Severity levels & Single & Three (Low/Mid/Hard) \\
Structured categorization & No & Four categories \\
\# base benchmarks & 1 & 6 \\
Capability diagnosis & Combined score & Per-benchmark \\
\bottomrule
\end{tabular}}
\end{table}

We include R-Bench-Dis as an additional evaluation set in our experiments (Table~\ref{tab:main}) to demonstrate that CLEAR generalizes to independently constructed degraded-image benchmarks beyond our own MMD-Bench.
\subsection{Base Benchmarks}

MMD-Bench is constructed by applying 16 real-world corruption types at three severity levels (Low, Mid, Hard) to six widely used multimodal benchmarks.
The six base benchmarks are selected to collectively cover a broad spectrum of multimodal understanding capabilities, from coarse-grained perception to fine-grained reasoning.

\textbf{MMBench}~\cite{mmbench} is a bilingual benchmark containing 2,974 multiple-choice questions that span 20 fine-grained ability dimensions organized into three hierarchical levels covering both perception and reasoning. It employs a CircularEval strategy that rotates the answer option order across multiple passes to reduce position bias, providing more reliable evaluation results than standard single-pass accuracy.

\textbf{MM-Vet}~\cite{mmvet} evaluates the integrated capabilities of multimodal models across six core vision-language dimensions: recognition, knowledge, OCR, spatial awareness, language generation, and math. It contains 218 open-ended questions over 200 images, and uses GPT-4 as an automated judge to score free-form responses, making it particularly suitable for evaluating complex answers that require multiple capabilities simultaneously.

\textbf{MMVP}~\cite{mmvp} is designed to probe visual perception failures that stem from CLIP-based vision encoders. It consists of 300 image pairs that appear similar to CLIP but differ in visually obvious ways to humans, paired with straightforward yes/no questions. MMVP is especially relevant to our study because the visual distinctions it tests are precisely the kind of fine-grained cues that degradation tends to destroy.

\textbf{CV-Bench}~\cite{cvbench} contains 2,638 manually inspected examples repurposed from classic computer vision benchmarks including ADE20K, COCO, and Omni3D. It assesses multimodal models on traditional vision tasks such as object detection, counting, and depth estimation within a VQA format, focusing on vision-centric spatial understanding that demands accurate low-level perception.

\textbf{MMStar}~\cite{mmstar} comprises 1,500 carefully curated samples designed to ensure visual dependency and minimal data leakage. It evaluates six core capabilities across 18 detailed axes, with each sample verified to be unanswerable without the visual input, making it a rigorous test of genuine multimodal reasoning rather than language-only shortcuts.

\textbf{RealWorldQA} consists of 764 images captured from real-world scenarios including driving scenes and everyday environments, each paired with a question about spatial relationships or scene understanding. It tests practical visual comprehension in naturalistic settings where image quality is inherently variable.

Table~\ref{tab:bench_stats} summarizes the key characteristics of each benchmark.

\begin{table}[h]
\centering
\caption{Base benchmarks used in MMD-Bench.}
\label{tab:bench_stats}
\resizebox{\linewidth}{!}{
\begin{tabular}{lcll}
\toprule
Benchmark & \# Samples & Primary Focus & Evaluation \\
\midrule
MMBench & 2,974 & Fine-grained multi-ability assessment & Accuracy (CircularEval) \\
MM-Vet & 218 & Integrated VL capability evaluation & GPT-4 scoring \\
MMVP & 300 & CLIP-blind visual perception & Accuracy \\
CV-Bench & 2,638 & Vision-centric spatial understanding & Accuracy \\
MMStar & 1,500 & Vision-indispensable reasoning & Accuracy \\
RealWorldQA & 764 & Real-world spatial comprehension & Accuracy \\
\bottomrule
\end{tabular}}
\end{table}

\begin{figure*}[h]
    \centering
    \includegraphics[width=\textwidth]{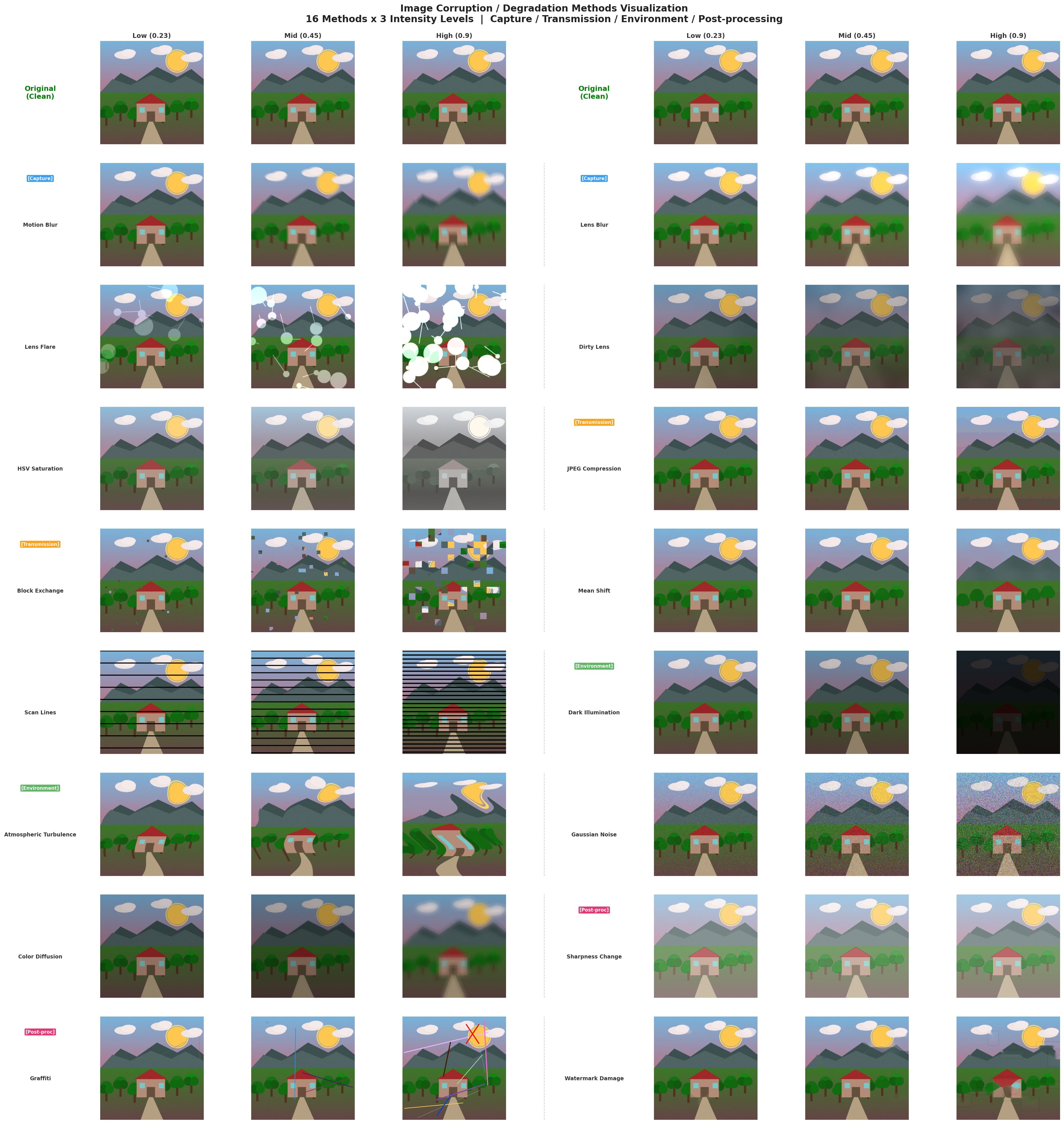}
    \caption{Visualization of all 16 corruption types at three severity levels. Each row shows one corruption type applied to the same source image at Low (left), Mid (center), and Hard (right) severity.}
    \label{fig:corruption_vis}
\end{figure*}

Table~\ref{tab:corruption_types} lists all 16 corruption types organized by category, and Figure~\ref{fig:corruption_vis} visualizes representative examples at each severity level.

\begin{table}[h]
\centering
\caption{The 16 corruption types in MMD-Bench, organized into four categories that reflect distinct real-world degradation sources.}
\label{tab:corruption_types}
\resizebox{\linewidth}{!}{
\begin{tabular}{lll}
\toprule
Category & Corruption Types & Real-world Source \\
\midrule
Capture & lens\_blur, lens\_flare, motion\_blur, & Camera hardware and \\
& dirty\_lens, hsv\_saturation & shooting conditions \\
\midrule
Transmission & jpeg\_compression, block\_exchange, & Lossy compression and \\
& mean\_shift, scan\_lines & bandwidth limitations \\
\midrule
Environmental & dark\_illumination, atmospheric\_turbulence, & Adverse lighting and \\
& gaussian\_noise, color\_diffusion & atmospheric conditions \\
\midrule
Post-processing & sharpness\_change, graffiti, & Downstream editing and \\
& watermark\_damage & overlay artifacts \\
\bottomrule
\end{tabular}}
\end{table}

For each corruption type, we define three severity levels that progressively increase the degradation strength.
Low degradation introduces mild perturbations that are noticeable but do not severely affect content understanding.
Mid degradation produces clearly visible artifacts that begin to impair fine-grained recognition.
Hard degradation substantially obscures visual details, making many low-level cues unrecoverable without additional information.
The severity parameters for each corruption type are calibrated empirically to ensure that these qualitative descriptions hold consistently across different image contents.

For each benchmark, all test images are corrupted with all 16 types at all three severity levels, yielding 48 degraded variants per image.
Evaluation follows the original benchmark protocols, with the only modification being the replacement of clean images with their degraded counterparts.
All evaluations are conducted using VLMEvalKit~\cite{duan2024vlmevalkit} to ensure reproducibility.
The reported score at each severity level is the average across all 16 corruption types, providing a comprehensive measure of robustness rather than sensitivity to any single degradation.

\section{Training Data Construction}
\label{app:data}

This section provides additional details on the construction of the degradation-aware SFT dataset and the GRPO training set described in Section~\ref{sec:sft}.

\subsection{Data Collection Pipeline}

We sample 48k image-question pairs from the LLaVA-OneVision~\cite{llavaonevision} instruction-tuning dataset, selecting samples that cover diverse visual domains including natural scenes, documents, charts, and everyday objects.
For each sampled image, we randomly select one of the 16 corruption types and one of the three severity levels to generate a degraded version.
We then query the base Bagel model with the degraded image and the associated question to determine whether the model can answer correctly.
Samples that the model answers correctly are assigned to the direct-answer pathway, while samples it fails on are assigned to the generate-then-answer pathway.
We balance the two pathways to a 1:1 ratio by subsampling the larger group.
The final 48k samples are split into two non-overlapping sets of 24k each, one for SFT and one for Interleaved GRPO.

\subsection{Reasoning Trace Generation}

For both pathway types, we use GPT-4.1~\cite{gpt41} to generate structured reasoning traces.
The generation prompt provides GPT-4.1 with the clean image, the degraded image, the question, and the ground-truth answer, and instructs it to produce a trace conforming to one of two patterns depending on the assigned pathway.
Figure~\ref{fig:trace_prompt} shows the prompt template for the generate-then-answer pathway.

\begin{figure}[h]
\small
\fbox{\parbox{0.92\linewidth}{
\textbf{Prompt Template for Generate-then-Answer Trace} \\[4pt]
You are an advanced AI training data generator. You will be given a degraded image, its clean version, a question, and the ground-truth answer. Your task is to synthesize a high-quality reasoning trace that follows the structure below. \\[4pt]
\textbf{Step A (Diagnosis):} Act as if seeing only the corrupted image. Describe the visual defects you observe. Hypothesize the degradation type. State that the quality is too poor to answer confidently and decide to invoke the restoration tool. Do NOT reveal or guess the answer at this stage. \\[4pt]
\textbf{Step B (Tool Trigger):} Output \texttt{<image\_restore>} on its own line. \\[4pt]
\textbf{Step C (Post-restoration Analysis):} Act as if you have received the restored image. Confirm that the previously observed artifacts are resolved. Locate the visual details that are now visible and relevant to answering the question. Connect these details to form a conclusion. \\[4pt]
\textbf{Step D (Answer):} Provide the final answer concisely, matching the ground-truth answer.
}}
\caption{Prompt template used to generate reasoning traces for the generate-then-answer pathway via GPT-4.1.}
\label{fig:trace_prompt}
\end{figure}

For the direct-answer pathway, Steps B and C are omitted.
The prompt instructs GPT-4.1 to diagnose the image condition, determine that the visual information is sufficient despite mild degradation, and proceed directly to reasoning and answering.

All generated traces are filtered against ground-truth answers.
Traces whose final answers do not match the ground truth are discarded and regenerated up to three times before the sample is dropped entirely.

\subsection{Dataset Statistics}

Table~\ref{tab:data_stats} summarizes the key statistics of the final SFT dataset.

\begin{table}[h]
\centering
\caption{SFT dataset statistics.}
\label{tab:data_stats}
\begin{tabular}{lc}
\toprule
Property & Value \\
\midrule
Total samples & 24,886 \\
Direct-answer samples & 12,267 \\
Generate-then-answer samples & 12,619 \\
Average trace length (direct) & 606 \\
Average trace length (generate) & 1080 \\
corruption types used & 16 \\
severity levels & 3 \\
Source dataset & LLaVA-OneVision \\
GRPO set (separate, non-overlapping) & 24,480 \\
\bottomrule
\end{tabular}
\end{table}

The degradation distribution in the SFT dataset is approximately uniform across the 16 corruption types and three severity levels, with minor imbalances arising from the pathway assignment process since harder corruptions are more likely to cause model failures and thus be assigned to the generate-then-answer pathway.

\section{System Prompt}
\label{app:prompt}

Figure~\ref{fig:system_prompt} shows the system prompt used throughout training (both SFT and Interleaved GRPO) and inference.
The prompt defines two reasoning scenarios: Scenario 1 for the generate-then-answer pathway when degradation obscures critical details, and Scenario 2 for direct answering when visual information is sufficient despite degradation.
It specifies the output structure with \texttt{<think>}, \texttt{<answer>}, and \texttt{<image\_restore>} tags, and requires the model to perform explicit image quality analysis before deciding whether to invoke generation.

\begin{figure*}[h]
\small
\fbox{\parbox{0.95\textwidth}{
\textbf{System Prompt} \\[4pt]
You are a specialized multimodal assistant. Your purpose is to solve visual question answering tasks by thinking step-by-step and utilizing an image restoration tool when necessary. \\[4pt]
\textbf{Skills.} You can trigger image restoration by generating the following special token sequence: \texttt{<image\_restore>}. This tool performs enhancement operations (e.g., deblurring, denoising) on the input image to reveal details that are currently obscured. \\[4pt]
\textbf{Instruction.} \\
(1) \textbf{Reasoning} (\texttt{<think>}): In each turn, you must start with a \texttt{<think>} tag. Inside, conduct a step-by-step reasoning process. Analyze image quality by identifying degradations (blur, noise, low resolution, etc.). Assess sufficiency by determining if the current image quality allows you to answer the question confidently. \\[2pt]
(2) \textbf{Tool Usage}: If the degradation prevents you from seeing critical details required for the answer, you MUST trigger the restoration tool by outputting \texttt{<image\_restore>}. If the answer is visible despite the degradation, do NOT use the tool. \\[2pt]
(3) \textbf{Answering} (\texttt{<answer>}): After reasoning (and potential restoration), provide your final response in the \texttt{<answer>} tag. The answer should be natural, concise, and direct. \\[2pt]
(4) \textbf{Format}: Keep your output compact. Avoid unnecessary newlines between tags. \\[4pt]
\textbf{Scenario 1} (restoration needed): \\
\texttt{<think>} The image is heavily blurred, making the text unreadable. I need to restore it to extract the information. \texttt{<\!/think> <image\_restore> <think>} The restored image is clear. The text says ``EXIT''. \texttt{<\!/think> <answer>} The text on the sign is ``EXIT''. \texttt{<\!/answer>} \\[4pt]
\textbf{Scenario 2} (direct answer): \\
\texttt{<think>} Although there is some noise, the red car is clearly visible in the foreground. \texttt{<\!/think> <answer>} The car is red. \texttt{<\!/answer>}
}}
\caption{System prompt used during SFT, Interleaved GRPO, and inference. The same prompt is shared across all stages without modification.}
\label{fig:system_prompt}
\end{figure*}

\section{Full Severity-Level Results}
\label{app:severity}

Table~\ref{tab:severity_full} reports the complete results of CLEAR-RL across all severity levels on each of the six MMD-Bench benchmarks.

\begin{table*}[h]
\centering
\caption{CLEAR-RL results across degradation severity levels on each MMD-Bench benchmark. AVG is computed over the six benchmarks.}
\label{tab:severity_full}
\resizebox{0.8\textwidth}{!}{
\begin{tabular}{l|cccccc|c}
\toprule
Level & MMBench & MM-Vet & MMVP & CV-Bench & MMStar & RealWorldQA & AVG \\
\midrule
Clean & 80.03 & 61.19 & 77.00 & 76.14 & 65.86 & 61.43 & 70.27 \\
Low & 79.41 & 57.93 & 75.33 & 75.86 & 64.40 & 63.39 & 69.39 \\
Mid & 78.48 & 55.27 & 75.00 & 75.17 & 64.60 & 61.69 & 68.37 \\
Hard & 72.52 & 51.97 & 71.33 & 72.25 & 60.67 & 61.05 & 64.97 \\
\bottomrule
\end{tabular}}
\end{table*}

Performance degrades gracefully from Clean to Hard, with the average dropping from 70.27 to 64.97 (a 5.30-point or 7.5\% relative decline).
The drop from Clean to Low is modest (0.88 points), indicating that CLEAR-RL handles mild degradation with minimal accuracy loss.
The steepest decline occurs between Mid and Hard (3.40 points), where severe corruptions begin to obscure critical visual details beyond what the generative pathway can fully recover.
Across benchmarks, MM-Vet shows the largest absolute drop from Clean to Hard (9.22 points), consistent with its reliance on integrated multi-cue reasoning where multiple visual details must be simultaneously recovered.
RealWorldQA is notably stable across severity levels (61.43 to 61.05), likely because its spatial reasoning questions depend more on scene layout than on fine texture details.

\section{Per-Corruption Analysis}
\label{app:percorruption}

To understand how CLEAR-RL performs across different degradation sources, we report accuracy for each of the 16 corruption types under Hard degradation, grouped by their four categories (Table~\ref{tab:corruption_types}).
Table~\ref{tab:per_corruption} compares Bagel with CLEAR-RL, averaged over the six MMD-Bench benchmarks.

\begin{table}[h]
\centering
\caption{Per-corruption accuracy under Hard degradation, averaged over six MMD-Bench benchmarks. Corruptions are grouped by category. $\Delta$ shows the improvement of CLEAR-RL over Bagel. Category averages are shown in italics.}
\label{tab:per_corruption}
\resizebox{0.85\linewidth}{!}{
\begin{tabular}{llccc}
\toprule
Category & Corruption & Bagel & CLEAR-RL & $\Delta$ \\
\midrule
\multirow{6}{*}{Capture}
& lens\_blur & 57.82 & 64.48 & +6.66 \\
& lens\_flare & 59.23 & 64.85 & +5.62 \\
& motion\_blur & 56.35 & 63.52 & +7.17 \\
& dirty\_lens & 58.94 & 64.25 & +5.31 \\
& hsv\_saturation & 59.51 & 64.63 & +5.12 \\
& \textit{Category avg.} & \textit{58.37} & \textit{64.35} & \textit{+5.98} \\
\midrule
\multirow{5}{*}{Transmission}
& jpeg\_compression & 61.18 & 66.12 & +4.94 \\
& block\_exchange & 58.73 & 64.31 & +5.58 \\
& mean\_shift & 61.82 & 66.93 & +5.11 \\
& scan\_lines & 60.14 & 65.62 & +5.48 \\
& \textit{Category avg.} & \textit{60.47} & \textit{65.75} & \textit{+5.28} \\
\midrule
\multirow{5}{*}{Environmental}
& dark\_illumination & 57.63 & 63.94 & +6.31 \\
& atmospheric\_turbulence & 58.35 & 64.12 & +5.77 \\
& gaussian\_noise & 59.84 & 66.25 & +6.41 \\
& color\_diffusion & 60.52 & 65.03 & +4.51 \\
& \textit{Category avg.} & \textit{59.09} & \textit{64.84} & \textit{+5.75} \\
\midrule
\multirow{4}{*}{Post-proc.}
& sharpness\_change & 62.31 & 67.08 & +4.77 \\
& graffiti & 59.84 & 63.42 & +3.58 \\
& watermark\_damage & 60.93 & 65.14 & +4.21 \\
& \textit{Category avg.} & \textit{61.03} & \textit{65.21} & \textit{+4.19} \\
\midrule
\multicolumn{2}{l}{\textbf{Overall}} & \textbf{59.57} & \textbf{64.97} & \textbf{+5.40} \\
\bottomrule
\end{tabular}}
\end{table}

CLEAR-RL improves over Bagel consistently across all 16 corruption types.
At the category level, capture degradations benefit the most (+5.98), as blur and flare destroy fine spatial structure that the generative pathway is particularly well-suited to recover.
Environmental degradations show the second largest gain (+5.75), where noise and poor illumination uniformly obscure texture and color details across the image.
Transmission degradations gain +5.28, with compression artifacts partially recoverable through the learned denoising trajectory.
Post-processing degradations benefit the least (+4.19), likely because corruptions such as graffiti and watermark overlay foreign content that is structurally different from natural image degradation, making them harder to address through the same generative process.

At the individual corruption level, motion\_blur (+7.17) and gaussian\_noise (+6.41) show the largest improvements.
Both corruptions uniformly degrade spatial structure across the entire image, creating exactly the type of low-level information loss that the generative pathway is designed to recover.
lens\_blur (+6.66) and dark\_illumination (+6.31) follow closely, as these similarly destroy fine detail in a spatially uniform manner.
In contrast, graffiti (+3.58) shows the smallest gain.
Unlike natural degradations that reduce image quality uniformly, graffiti overlays spatially localized foreign content onto the image, and the denoising process must distinguish between original content and overlaid artifacts, a fundamentally harder task than recovering information that has been blurred or noised.
color\_diffusion (+4.51) and sharpness\_change (+4.77) also show moderate gains, as these corruptions alter global image properties in ways that partially preserve the structural cues the understanding pathway can still exploit, reducing the marginal benefit of generation.

This fine-grained analysis confirms that the generate-then-answer strategy is broadly effective rather than specialized to any particular degradation type, while also revealing that the generative pathway is most beneficial when degradation uniformly destroys spatial structure and least beneficial when corruptions introduce foreign visual content.

\section{Inference Latency}
\label{app:latency}

The main text (Figure~\ref{fig:trigger_and_cost}) shows that inference time closely tracks the generation triggering rate.
Table~\ref{tab:latency} reports the evaluation time of CLEAR-RL on each benchmark across degradation levels, measured on a single NVIDIA A100 80GB GPU.

\begin{table}[h]
\centering
\caption{CLEAR-RL evaluation time across degradation levels on each benchmark. R-Bench-Dis is evaluated only once as it contains pre-degraded images at a single severity level.}
\label{tab:latency}
\resizebox{0.7\linewidth}{!}{
\begin{tabular}{lccc}
\toprule
Benchmark & High & Mid & Low \\
\midrule
MMBench & 1h 35m & 1h 01m & 54m \\
MM-Vet & 9m & 7m & 7m \\
MMVP & 8m & 6m & 5m \\
CV-Bench & 31m & 19m & 15m \\
MMStar & 43m & 31m & 28m \\
RealWorldQA & 27m & 17m & 13m \\
\midrule
R-Bench-Dis & \multicolumn{3}{c}{12m} \\
\midrule
Total & 3h 35m & 2h 23m & 2h 03m \\
\bottomrule
\end{tabular}}
\end{table}

Inference time increases monotonically with degradation severity across all benchmarks, driven by the higher generation triggering rate.
Under Low degradation, the average triggering rate is only 5.2\% and total evaluation time across the six MMD-Bench benchmarks is 2 hours 3 minutes.
Under High degradation, the triggering rate rises to 36.4\% and total time increases to 3 hours 35 minutes, a 74\% increase.
The per-benchmark pattern is consistent: MMBench, with the largest sample count (2,974), shows the largest absolute time increase (54m $\to$ 1h 35m), while smaller benchmarks like MMVP (300 samples) show proportionally smaller increases (5m $\to$ 8m).

The time difference between severity levels is entirely attributable to the adaptive generation policy.
When generation is not triggered, the model performs only text reasoning with overhead comparable to Text-only CoT.
When generation is triggered, the 30-step denoising process adds a fixed per-sample cost.
The adaptive policy thus concentrates computational resources on inputs where generation yields the largest accuracy benefit, keeping overhead moderate under mild degradation while accepting the additional cost under severe degradation where the accuracy gains justify it.

\section{Hyperparameter Sensitivity}
\label{app:hyperparameter}

We analyze the sensitivity of CLEAR-RL to key hyperparameters by varying one parameter at a time while keeping all others at their default values.
All experiments are evaluated on the six MMD-Bench benchmarks under both Clean and Hard degradation.

\subsection{Flow-GRPO Loss Weight $\lambda$}

The weight $\lambda$ in $\mathcal{L}_\text{Interleaved} = \mathcal{L}_\text{GRPO} + \lambda\,\mathcal{L}_\text{Flow-GRPO}$ controls the relative contribution of the image generation objective.
Table~\ref{tab:lambda} reports the results across different values.

\begin{table}[h]
\centering
\caption{Effect of Flow-GRPO loss weight $\lambda$ on accuracy (6-bench average).}
\label{tab:lambda}
\resizebox{0.55\linewidth}{!}{
\begin{tabular}{lcc}
\toprule
$\lambda$ & Clean & Hard \\
\midrule
0.0 & 69.85 & 63.52 \\
0.1 & 70.06 & 64.31 \\
0.3 (default) & \textbf{70.27} & \textbf{64.97} \\
0.5 & 70.18 & 64.72 \\
1.0 & 69.71 & 63.89 \\
\bottomrule
\end{tabular}}
\end{table}

Setting $\lambda = 0$ reduces Interleaved GRPO to text-only GRPO, where the generation process receives no direct optimization signal.
This still outperforms SFT (63.04 Hard) because the text-side GRPO improves reasoning, but the generative pathway is not optimized and remains at its SFT initialization.
Performance improves as $\lambda$ increases from 0 to 0.3, confirming that coupling image generation to the reward signal is beneficial.
Beyond 0.3, performance begins to decline: at $\lambda = 1.0$, the image-side gradients become too dominant relative to the text-side, slightly destabilizing the text reasoning process.
The default value of 0.3 provides the best balance, and the method is reasonably robust within the range 0.1 to 0.5.

\subsection{Reward Weights}

Table~\ref{tab:reward_weights} examines the sensitivity to the three reward components by varying $w_\text{acc}$, $w_\text{fmt}$, and $w_\text{dec}$.

\begin{table}[h]
\centering
\caption{Effect of reward weight configurations on accuracy (6-bench average).}
\label{tab:reward_weights}
\resizebox{0.7\linewidth}{!}{
\begin{tabular}{ccccc}
\toprule
$w_\text{acc}$ & $w_\text{fmt}$ & $w_\text{dec}$ & Clean & Hard \\
\midrule
1.0 & 0.0 & 0.0 & 69.92 & 64.18 \\
0.85 & 0.1 & 0.05 & 70.13 & 64.55 \\
0.75 & 0.1 & 0.15 (default) & \textbf{70.27} & \textbf{64.97} \\
0.65 & 0.1 & 0.25 & 70.08 & 64.61 \\
0.60 & 0.15 & 0.25 & 69.83 & 64.24 \\
\bottomrule
\end{tabular}}
\end{table}

Using only the accuracy reward ($w_\text{dec} = 0$) yields a 0.79-point drop on Hard compared to the default, because without the decision reward the model lacks a direct signal for learning when to trigger generation.
In this configuration the model tends to either over-generate on easy inputs or under-generate on hard inputs, as both behaviors can occasionally lead to correct answers and thus receive similar accuracy rewards.
Increasing $w_\text{dec}$ to 0.15 provides the strongest performance by sharpening the generation decision.
However, pushing $w_\text{dec}$ further to 0.25 shifts too much focus toward the binary generation decision at the expense of answer quality, leading to a slight decline.
The format reward $R_\text{fmt}$ is necessary to prevent degenerate outputs in early training but has limited influence on final performance as long as it is present.

\subsection{Denoising Steps}

Table~\ref{tab:denoise_steps} varies the number of denoising steps used during both training and inference.

\begin{table}[h]
\centering
\caption{Effect of denoising steps on accuracy (6-bench Hard average) and per-sample denoising time (measured on samples that trigger generation).}
\label{tab:denoise_steps}
\resizebox{0.6\linewidth}{!}{
\begin{tabular}{lcc}
\toprule
Steps & Hard & Denoising Time \\
\midrule
10 & 63.68 & 1.8s \\
20 & 64.53 & 3.5s \\
30 (default) & \textbf{64.97} & 5.2s \\
50 & 65.04 & 8.7s \\
\bottomrule
\end{tabular}}
\end{table}

Accuracy improves from 10 to 30 steps as the denoising process has more iterations to recover fine details from the degraded input.
The gain from 30 to 50 steps is marginal (0.07 points) while denoising time increases by 67\%, making 30 steps a favorable trade-off between accuracy and efficiency.
At 10 steps the denoising process is too coarse to recover the structural detail needed for reasoning, resulting in a 1.29-point drop compared to the default.
The per-sample denoising time scales approximately linearly with the number of steps, confirming that the computational cost is predictable and controllable.

\section{Reward Design Details}
\label{app:reward}

This section provides additional details on the three reward components used in Interleaved GRPO.

\subsection{Accuracy Reward $R_\text{acc}$}

The accuracy reward is computed by prompting GPT-4.1-mini to compare the model's answer with the ground-truth answer on a scale from 0 to 1.
Figure~\ref{fig:judge_prompt} shows the prompt template.
The judge is instructed to focus on semantic correctness rather than surface-level string matching, allowing partial credit for answers that capture the correct concept but differ in phrasing or formatting.

\begin{figure}[h]
\small
\fbox{\parbox{0.92\linewidth}{
\textbf{LLM-as-Judge Prompt} \\[4pt]
You are an impartial judge evaluating the correctness of an AI assistant's answer to a visual question. \\[2pt]
\textbf{Ground Truth:} \{ground\_truth\} \\
\textbf{Model Answer:} \{model\_answer\} \\[2pt]
Rate the correctness of the model answer on a scale from 0.0 to 1.0, where 0.0 means completely wrong and 1.0 means perfectly correct. Focus on semantic meaning rather than exact wording. Award partial credit if the answer captures the correct concept but includes minor errors in phrasing, formatting, or specificity. \\[2pt]
Output only a single number between 0.0 and 1.0.
}}
\caption{Prompt template for the LLM-as-judge accuracy evaluation.}
\label{fig:judge_prompt}
\end{figure}

\subsection{Format Reward $R_\text{fmt}$}

The format reward is a binary signal that checks whether the model output conforms to the expected structure.
An output receives $R_\text{fmt} = 1$ if it contains properly formed \texttt{<think>} and \texttt{<answer>} blocks and follows one of the two valid patterns defined in the system prompt.
Otherwise $R_\text{fmt} = 0$.
This reward is intentionally simple, serving only to prevent degenerate outputs during early GRPO training.

\subsection{Decision Reward $R_\text{dec}$}

The decision reward evaluates the generation decision retrospectively based on the accuracy score.
Let $g \in \{0, 1\}$ indicate whether the model triggered generation, and let $c \in \{0, 1\}$ indicate whether the answer is correct (defined as $R_\text{acc} > 0.5$).
The four cases are:

\begin{table}[h]
\centering
\caption{Decision reward $R_\text{dec}$ for the four possible outcomes.}
\label{tab:rdec}
\resizebox{0.7\linewidth}{!}{
\begin{tabular}{cccc}
\toprule
Generated ($g$) & Correct ($c$) & $R_\text{dec}$ & Interpretation \\
\midrule
1 & 1 & $+1$ & Generation helped \\
0 & 0 & $-1$ & Should have generated \\
1 & 0 & $0$ & Neutral \\
0 & 1 & $0$ & Neutral \\
\bottomrule
\end{tabular}}
\end{table}

The asymmetric design rewards successful generation and penalizes missed opportunities, while treating the remaining two cases as neutral.
Incorrect answers after generation are not penalized because the model correctly identified the need for generation even though the restored information was insufficient.
Correct answers without generation are not rewarded because no generation decision was needed.
This design allows the adaptive generation policy to emerge naturally from the reward signal without requiring an explicit decision classifier.

\section{Qualitative Results}
\label{app:qualitative}

This section presents qualitative examples of CLEAR-RL's reasoning behavior across different degradation types and scenarios.

\subsection{Reasoning Trace Examples}

We first show complete reasoning trajectories to illustrate the two reasoning patterns and a representative failure case.

\begin{figure}[h]
\small
\fbox{\parbox{0.92\linewidth}{
\textbf{Question:} What text is written on the sign? \\
\textbf{Degradation:} Motion blur (Hard) \\[4pt]
\texttt{<think>} The image suffers from severe motion blur. The sign in the center of the image has text on it, but the blur makes the characters unreadable. I cannot determine the text with confidence from the current image quality. I will request image restoration to recover the text details. \texttt{</think>} \\[2pt]
\texttt{<image\_restore>} \\[2pt]
\texttt{<think>} The restored image is much clearer. I can now see the sign in the center of the image. The text on the sign reads ``NO PARKING''. The letters are white on a red background. \texttt{</think>} \\[2pt]
\texttt{<answer>} The sign says ``NO PARKING''. \texttt{</answer>}
}}
\caption{Generate-then-answer trajectory. The model diagnoses motion blur, triggers restoration, and extracts the recovered text.}
\label{fig:trace_generate}
\end{figure}

\begin{figure}[h]
\small
\fbox{\parbox{0.92\linewidth}{
\textbf{Question:} How many people are in the image? \\
\textbf{Degradation:} Gaussian noise (Hard) \\[4pt]
\texttt{<think>} The image has noticeable Gaussian noise, but the overall scene is still discernible. I can see three people standing near a table in what appears to be a kitchen. The noise affects fine textures but the human figures are clearly distinguishable by their outlines and positioning. I can answer this question without restoration. \texttt{</think>} \\[2pt]
\texttt{<answer>} There are three people in the image. \texttt{</answer>}
}}
\caption{Direct-answer trajectory. The model judges that visual information is sufficient despite noise and answers without generation.}
\label{fig:trace_direct}
\end{figure}

\begin{figure}[h]
\small
\fbox{\parbox{0.92\linewidth}{
\textbf{Question:} What brand is the laptop on the desk? \\
\textbf{Degradation:} JPEG compression (Hard) \\[4pt]
\texttt{<think>} The image is heavily compressed with visible blocking artifacts. I can see a laptop on a wooden desk, but the brand logo on the laptop lid is very small and the compression has destroyed the fine detail in that region. I will request restoration. \texttt{</think>} \\[2pt]
\texttt{<image\_restore>} \\[2pt]
\texttt{<think>} The restored image has improved overall clarity, but the brand logo remains too small and the compression artifacts in that specific region were too severe to fully recover. The shape of the logo suggests it could be Dell or HP, but I cannot determine this with certainty. Based on the overall shape, I will provide my best estimate. \texttt{</think>} \\[2pt]
\texttt{<answer>} Dell \texttt{</answer>} \\[2pt]
\textit{Ground truth: HP}
}}
\caption{Failure case. Generation improves overall quality but cannot recover the fine detail needed for the answer. The model identifies its uncertainty but guesses incorrectly.}
\label{fig:trace_failure}
\end{figure}

The failure case illustrates a limitation of the current approach.
When the critical visual evidence occupies a very small region and is severely corrupted, the 30-step denoising process may not recover sufficient detail for correct identification, even though the overall image quality improves.
This suggests that future work on region-aware or adaptive-resolution generation could further improve performance on such cases.

\subsection{Additional Visual Examples}

\begin{figure*}[h]
    \centering
    \includegraphics[width=0.8\textwidth]{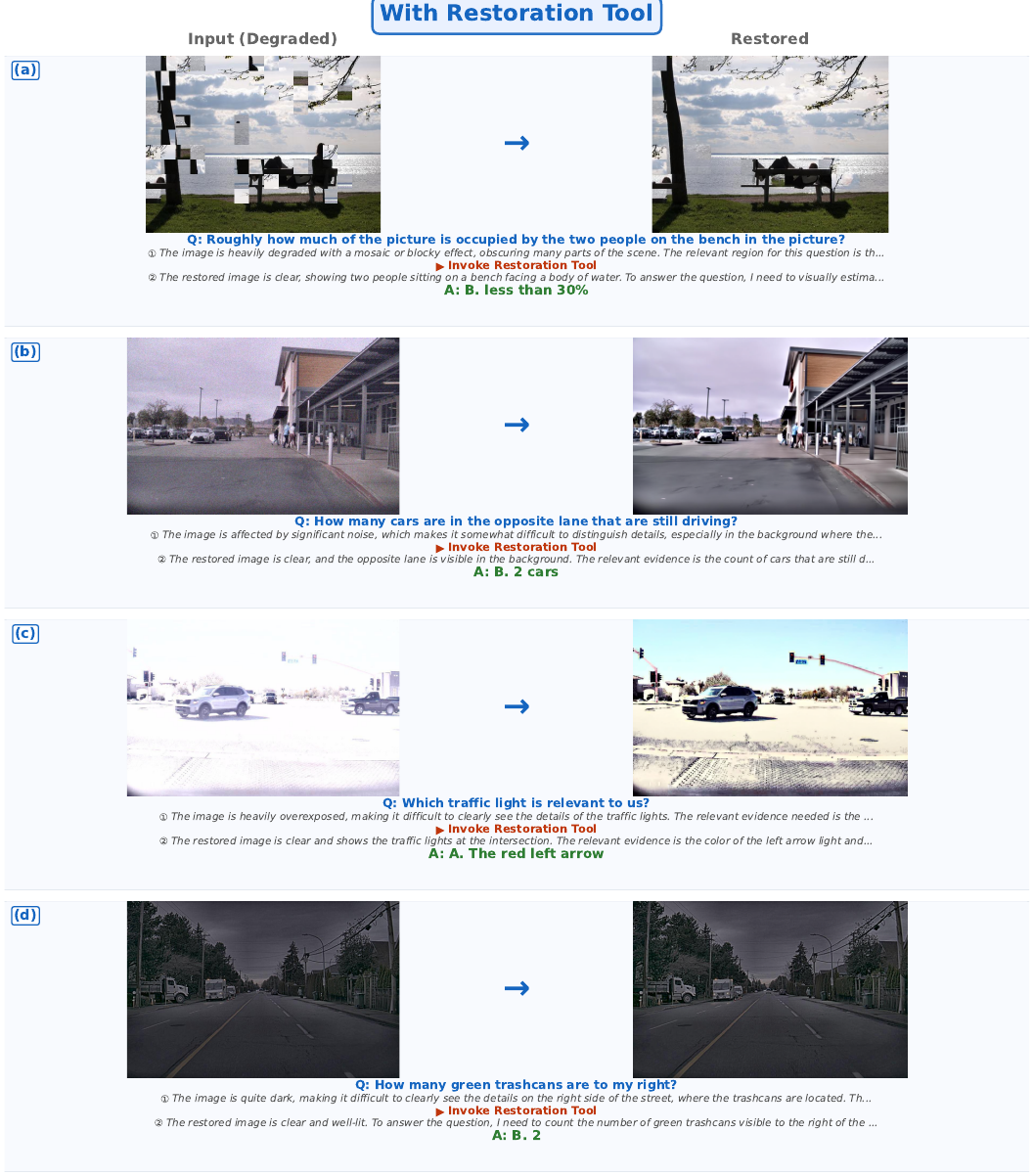}
    \caption{Additional qualitative examples across different degradation types.}
    \label{fig:additional_qual}
\end{figure*}

Figure~\ref{fig:additional_qual} presents additional examples across different degradation types and benchmarks.
Each example shows the degraded input, the intermediate visual state produced by CLEAR-RL (when generation is triggered), and the model's final answer alongside the ground truth.
These examples further illustrate two key behaviors.
First, the adaptive generation policy consistently triggers generation for severe degradations that obscure critical visual details while skipping generation for mild degradations where the understanding pathway alone is sufficient.
Second, the intermediate visual states recover task-relevant structure such as text, object boundaries, and spatial layout, consistent with the finding that task-driven optimization prioritizes reasoning utility.

\end{document}